\documentclass[runningheads]{llncs}

\usepackage{eccv}

\usepackage{eccvabbrv}

\usepackage{graphicx}
\usepackage{booktabs}
\usepackage{multirow}
\usepackage{multicol}

\usepackage{enumitem}
\setlist{topsep=0pt, leftmargin=*}

\usepackage[accsupp]{axessibility}  %

\usepackage{tikz,subcaption}
\usetikzlibrary{spy,backgrounds} 
\usepackage{xcolor}
\usepackage{amsmath}
\usepackage[ruled]{algorithm2e}
\usepackage{algpseudocode}
\usepackage{nicefrac}
\usepackage{indentfirst}
\usepackage{tabularx}
\usepackage{color}
\usepackage{colortbl}
\usepackage{graphicx}
\usepackage{lipsum}
\usepackage{wrapfig}
\usepackage{graphicx}
\usepackage{caption}
\usepackage{subcaption}
\usepackage{tikz}
\usepackage{pgfplots}
\usetikzlibrary{spy,calc}
\usepackage{hyperref}
\newif\ifblackandwhitecycle
\gdef\patternnumber{0}
\pgfkeys{/tikz/.cd,
    zoombox paths/.style={
        draw=orange,
        very thick
    },
    black and white/.is choice,
    black and white/.default=static,
    black and white/static/.style={ 
        draw=white,   
        zoombox paths/.append style={
            draw=white,
            postaction={
                draw=black,
                loosely dashed
            }
        }
    },
    black and white/static/.code={
        \gdef\patternnumber{1}
    },
    black and white/cycle/.code={
        \blackandwhitecycletrue
        \gdef\patternnumber{1}
    },
    black and white pattern/.is choice,
    black and white pattern/0/.style={},
    black and white pattern/1/.style={    
            draw=white,
            postaction={
                draw=black,
                dash pattern=on 2pt off 2pt
            }
    },
    black and white pattern/2/.style={    
            draw=white,
            postaction={
                draw=black,
                dash pattern=on 4pt off 4pt
            }
    },
    black and white pattern/3/.style={    
            draw=white,
            postaction={
                draw=black,
                dash pattern=on 4pt off 4pt on 1pt off 4pt
            }
    },
    black and white pattern/4/.style={    
            draw=white,
            postaction={
                draw=black,
                dash pattern=on 4pt off 2pt on 2 pt off 2pt on 2 pt off 2pt
            }
    },
    zoomboxarray inner gap/.initial=5pt,
    zoomboxarray columns/.initial=2,
    zoomboxarray rows/.initial=2,
    subfigurename/.initial={},
    figurename/.initial={zoombox},
    zoomboxarray/.style={
        execute at begin picture={
            \begin{scope}[
                spy using outlines={%
                    zoombox paths,
                    width=\imagewidth / \pgfkeysvalueof{/tikz/zoomboxarray columns} - (\pgfkeysvalueof{/tikz/zoomboxarray columns} - 1) / \pgfkeysvalueof{/tikz/zoomboxarray columns} * \pgfkeysvalueof{/tikz/zoomboxarray inner gap} -\pgflinewidth,
                    height=\imageheight / \pgfkeysvalueof{/tikz/zoomboxarray rows} - (\pgfkeysvalueof{/tikz/zoomboxarray rows} - 1) / \pgfkeysvalueof{/tikz/zoomboxarray rows} * \pgfkeysvalueof{/tikz/zoomboxarray inner gap}-\pgflinewidth,
                    magnification=3,
                    every spy on node/.style={
                        zoombox paths
                    },
                    every spy in node/.style={
                        zoombox paths
                    }
                }
            ]
        },
        execute at end picture={
            \end{scope}
            \node at (image.north) [anchor=north,inner sep=0pt] {\subcaptionbox{\label{\pgfkeysvalueof{/tikz/figurename}-image}}{\phantomimage}};
            \node at (zoomboxes container.north) [anchor=north,inner sep=0pt] {\subcaptionbox{\label{\pgfkeysvalueof{/tikz/figurename}-zoom}}{\phantomimage}};
     \gdef\patternnumber{0}
        },
        spymargin/.initial=0.5em,
        zoomboxes xshift/.initial=1,
        zoomboxes right/.code=\pgfkeys{/tikz/zoomboxes xshift=1},
        zoomboxes left/.code=\pgfkeys{/tikz/zoomboxes xshift=-1},
        zoomboxes yshift/.initial=0,
        zoomboxes above/.code={
            \pgfkeys{/tikz/zoomboxes yshift=1},
            \pgfkeys{/tikz/zoomboxes xshift=0}
        },
        zoomboxes below/.code={
            \pgfkeys{/tikz/zoomboxes yshift=-1},
            \pgfkeys{/tikz/zoomboxes xshift=0}
        },
        caption margin/.initial=4ex,
    },
    adjust caption spacing/.code={},
    image container/.style={
        inner sep=0pt,
        at=(image.north),
        anchor=north,
        adjust caption spacing
    },
    zoomboxes container/.style={
        inner sep=0pt,
        at=(image.north),
        anchor=north,
        name=zoomboxes container,
        xshift=\pgfkeysvalueof{/tikz/zoomboxes xshift}*(\imagewidth+\pgfkeysvalueof{/tikz/spymargin}),
        yshift=\pgfkeysvalueof{/tikz/zoomboxes yshift}*(\imageheight+\pgfkeysvalueof{/tikz/spymargin}+\pgfkeysvalueof{/tikz/caption margin}),
        adjust caption spacing,
    },
    calculate dimensions/.code={
        \pgfpointdiff{\pgfpointanchor{image}{south west} }{\pgfpointanchor{image}{north east} }
        \pgfgetlastxy{\imagewidth}{\imageheight}
        \global\let\imagewidth=\imagewidth
        \global\let\imageheight=\imageheight
        \gdef\columncount{1}
        \gdef\rowcount{1}
        
    },
    image node/.style={
        inner sep=0pt,
        name=image,
        anchor=south west,
        append after command={
            [calculate dimensions]
            node [image container,subfigurename=\pgfkeysvalueof{/tikz/figurename}-image] {\phantomimage}
            node [zoomboxes container,subfigurename=\pgfkeysvalueof{/tikz/figurename}-zoom] {\phantomimage}
        }
    },
    color code/.style={
        zoombox paths/.append style={draw=#1}
    },
    connect zoomboxes/.style={
    spy connection path={\draw[draw=none,zoombox paths] (tikzspyonnode) -- (tikzspyinnode);}
    },
    help grid code/.code={
        \begin{scope}[
                x={(image.south east)},
                y={(image.north west)},
                font=\footnotesize,
                help lines,
                overlay
            ]
            \foreach \x in {0,1,...,9} { 
                \draw(\x/10,0) -- (\x/10,1);
                \node [anchor=north] at (\x/10,0) {0.\x};
            }
            \foreach \y in {0,1,...,9} {
                \draw(0,\y/10) -- (1,\y/10);                        \node [anchor=east] at (0,\y/10) {0.\y};
            }
        \end{scope}    
    },
    help grid/.style={
        append after command={
            [help grid code]
        }
    },
}
\newcommand\phantomimage{%
    \phantom{%
        \rule{\imagewidth}{\imageheight}%
    }%
}
\newcommand\zoombox[2][]{
    \begin{scope}[zoombox paths]
        \pgfmathsetmacro\xpos{
            (\columncount-1)*(\imagewidth / \pgfkeysvalueof{/tikz/zoomboxarray columns} + \pgfkeysvalueof{/tikz/zoomboxarray inner gap} / \pgfkeysvalueof{/tikz/zoomboxarray columns} ) + \pgflinewidth
        }
        \pgfmathsetmacro\ypos{
            (\rowcount-1)*( \imageheight / \pgfkeysvalueof{/tikz/zoomboxarray rows} + \pgfkeysvalueof{/tikz/zoomboxarray inner gap} / \pgfkeysvalueof{/tikz/zoomboxarray rows} ) + 0.5*\pgflinewidth
        }
        \edef\dospy{\noexpand\spy [
            #1,
            zoombox paths/.append style={
                black and white pattern=\patternnumber
            },
            every spy on node/.append style={#1},
            x=\imagewidth,
            y=\imageheight
        ] on (#2) in node [anchor=north west] at ($(zoomboxes container.north west)+(\xpos pt,-\ypos pt)$);}
        \dospy
        \pgfmathtruncatemacro\pgfmathresult{ifthenelse(\columncount==\pgfkeysvalueof{/tikz/zoomboxarray columns},\rowcount+1,\rowcount)}
        \global\let\rowcount=\pgfmathresult
        \pgfmathtruncatemacro\pgfmathresult{ifthenelse(\columncount==\pgfkeysvalueof{/tikz/zoomboxarray columns},1,\columncount+1)}
        \global\let\columncount=\pgfmathresult
        \ifblackandwhitecycle
            \pgfmathtruncatemacro{\newpatternnumber}{\patternnumber+1}
            \global\edef\patternnumber{\newpatternnumber}
        \fi
    \end{scope}
}

\usepackage{multirow}
\usepackage[final]{changes}     %

\def\bx{{\bf x}}
\def\by{{\bf y}}
\def\bxa{{\bf x}^{acq}_i}
\def\bya{{\bf y}^{acq}_i}
\def\bw{{\bf w}}

\def\bH{{\mathbf H}}
\DeclareMathOperator{\tr}{tr}

\DeclareMathOperator*{\argmax}{arg\,max}

\DeclareMathOperator{\diag}{diag}

\usepackage{hyperref}

\usepackage{orcidlink}

\begin{document}

\title{FisherRF: Active View Selection and Mapping with Radiance Fields using Fisher Information}

\titlerunning{FisherRF}

\author{Wen Jiang\orcidlink{0000-0001-8730-7678}\inst{1} \and
Boshu Lei\orcidlink{0009-0000-1153-1702}\inst{1} \and
Kostas Daniilidis\orcidlink{0000-0003-0498-0758}\inst{1, 2}}

\authorrunning{W.~Jiang, W.~Lei, K.~Daniilidis}

\institute{$^1$University of Pennsylvania $\qquad$ $^2$Archimedes, Athena RC}

\maketitle

\begin{abstract}
This study addresses the challenging problem of active view selection and uncertainty quantification within the domain of Radiance Fields. Neural Radiance Fields (NeRF) have greatly advanced image rendering and reconstruction, but the cost of acquiring images poses the need to select the most informative viewpoints efficiently.
Existing approaches depend on modifying the model architecture or hypothetical perturbation field to indirectly approximate the model uncertainty. However, selecting views from indirect approximation does not guarantee optimal information gain for the model.
By leveraging Fisher Information, we directly quantify observed information on the parameters of Radiance Fields and select candidate views by maximizing the Expected Information Gain~(EIG).
Our method achieves state-of-the-art results on multiple tasks, including view selection, active mapping, and uncertainty quantification, demonstrating its potential to advance the field of Radiance Fields. 
\end{abstract}
    
\section{Introduction}
\label{sec:intro}
\begin{figure}
    \centering
 \includegraphics[width=\textwidth]{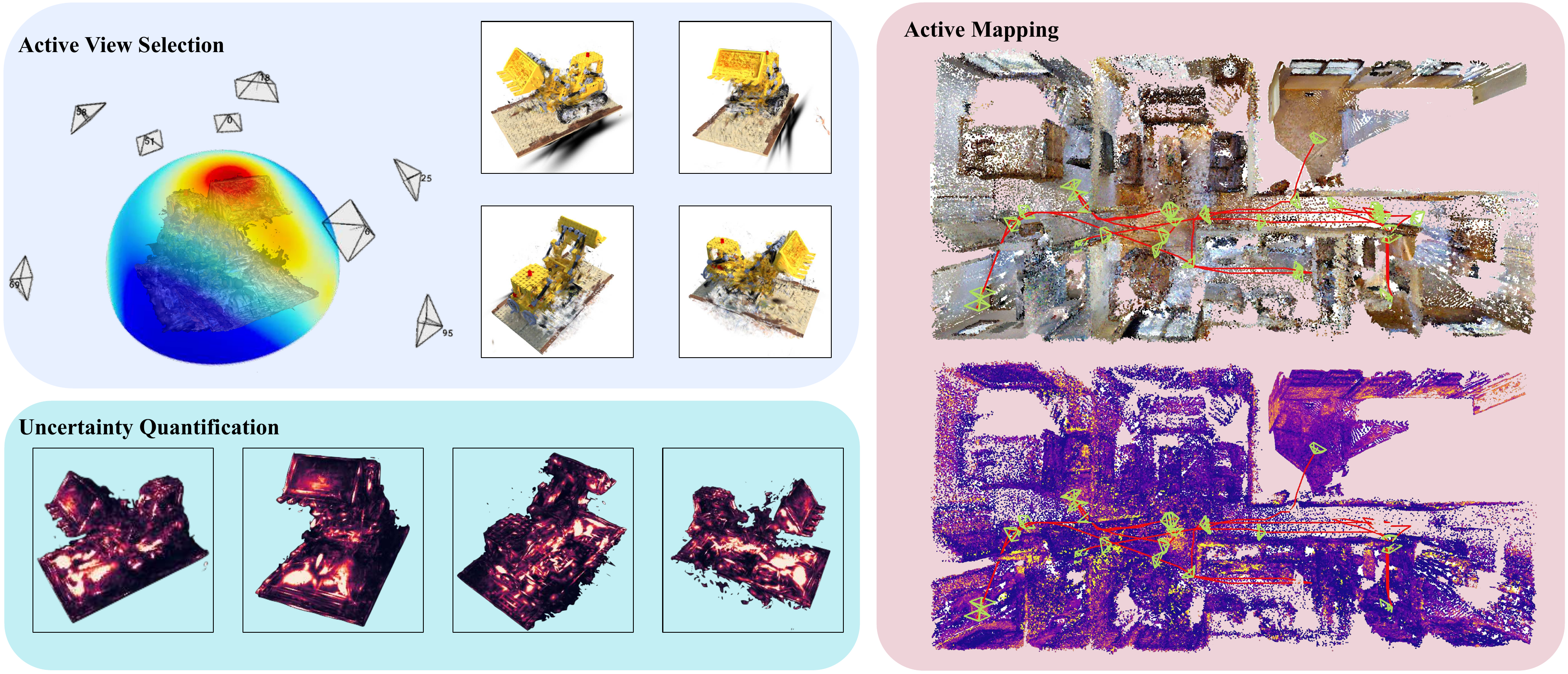}
    \caption{{\bf A brief overview of our method} Given a Radiance Field that was trained with a limited number of views, our method could find the next best view that could maximize information gain by computing the Fisher Information of the radiance field. We illustrate the Information Gain as a heat map on the viewing sphere and show four of the candidate views. Our model can quantify pixel-wise uncertainty, visualized at the bottom left,  by examining the Fisher Information on the related parameters of the ray. Our algorithm can be also adapted into an active mapping system, as showcased at the right, that could actively explore and reconstruct the environment.}
    \label{fig:teaser}
\end{figure}

\replaced{Neural Radiance Fields brought back image rendering and reconstruction from multiple views to the center of attention in the field of computer vision. Novel volumetric representations of radiance fields and differentiable volumetric rendering enabled unprecedented advances in image-based rendering of complex scenes both in terms of perceptual quality and speed.}
{Neural Radiance Fields (NeRF) have recently garnered significant attention in the field of Computer Vision and have seen enhancements for faster rendering and training, larger-scale reconstruction, dynamic modeling, and generative capabilities.} 

Recently, 3D Gaussian Splatting~\cite{kerbl3Dgaussians} has demonstrated distinct advantages in real-time rendering and explicit point-based parameterizations without neural representations.
However, to achieve satisfactory rendering quality, numerous viewpoints are required to train a radiance field, especially in large-scale scenarios. 
It is crucial to establish a criterion for selecting information-maximizing views before obtaining image observations at those locations.
Quantifying the observed information of a Radiance Field model is challenging, given that Radiance Field models are typically regression-based and scene-specific. The challenge intensifies when we aim to leverage quantified observed information for active view selection and mapping, especially when the selection candidates are only $SE(3)$ camera poses for acquiring new observations, a.k.a capturing new images.

Previous approaches select viewpoints by quantifying the uncertainty in Radiance Fields. They can be broadly categorized into two groups: variational white-box models and black-box models. White-box models integrate conventional NeRF architectures with Bayesian models, such as reparameterization~\cite{pan2022activenerf,martinbrualla2020nerfw,Ran2023neurar} and Normalizing Flows~\cite{CF-NeRF}. Black-box methods, on the other hand, do not modify the existing model architecture but seek to quantify predictive uncertainty by examining the distribution of predicted outcomes~\cite{zhan2022activermap,lee2022uncertainty, yan2023active-neural-mapping,pan2022activenerf}.
White-box models depend on specific model architectures and are often characterized by slower training times due to the challenges associated with probabilistic learning. Conversely, existing black-box models either focus solely on studying the uncertainty on NeRF-style query points through network prediction \cite{pan2022activenerf} or hypothetical perturbation field \cite{goli2023}, or rely on Monte Carlo sampling techniques~\cite{yan2023active-neural-mapping} to quantify the uncertainty at the candidate views.

In this study, our primary objective is to quantify the observed information of a Radiance Field model and utilize it to select the optimal view with the highest information gain for downstream tasks, Fig. \ref{fig:teaser}. To achieve this, we propose using Fisher Information, which represents the expectation of observation information. This quantity is directly linked to the second-order derivatives or Hessian matrix of the loss function involved in optimizing Radiance Field models. 
Importantly, the Hessian of the objective function in volumetric rendering is independent of ground truth data or the actual image measurement. This property allows us to compute the information gain between the training dataset and the candidate viewpoint pool using only the camera parameters of the candidate views. This capability facilitates an efficient next-best-view selection algorithm.
In addition to the formulation of Information Gain, the Hessian of the loss function has an intuitive interpretation: the perturbation at flat minima of the loss function. From an intuitive standpoint, one can perceive the Fisher Information matrix as a metric of the curvature of the log-likelihood function at specific parameter instantiations denoted as $\bw^*$.
Lower Fisher Information suggests that the log-likelihood function exhibits a flatter profile around $\bw^*$, implying that the loss is less prone to changes when $\bw^*$ is perturbed. The flat minimum interpretation has attracted substantial attention and research within various domains of machine learning \cite{Hinton1993, kirsch2022unifying, HochreiterFlat, SmithSGD, macdonald2023progressive}.
Moreover, as we have estimated the Fisher information for the model's parameters, which correspond to specific 3D locations in recent Radiance Field models such as 3D Gaussian Splatting~\cite{kerbl3Dgaussians} and Plenoxels~\cite{plenoxels}, we can derive pixel-wise uncertainty in the model's predictions by examing the Fisher Information on parameters that contribute to the prediction for each pixel.

We implement the computation of Fisher Information on top of two types of Radiance Field models: point-based 3D Gaussian Splatting~\cite{kerbl3Dgaussians} and Plenoxels~\cite{plenoxels}. 
In 3D Gaussian Splatting, we compute the Hessians of the parameters on each 3D Gaussian with respect to the negative log-likelihood. 
As Fisher Information is additive, we further extend our view selection algorithm into batch selection and path selection, which are closer to real-world applications.
To make the computation of information gain tractable, we use an approximation of the decrease in entropy by its upper bound, which is the trace of the product of the Hessian of the candidate view times the Hessian with respect to all previous views. These matrices are highly sparse due to the disentangling of scene parameters with respect to disparate rays. The sparsity allows us a computation of the matrix trace above that is as cost-effective as back-propagation, enabling us to evaluate views at 70 fps when we use 3D Gaussian Splatting. 
We carried out extensive evaluations in different benchmarks, including active view selection, active mapping, and uncertainty quantification. 
The quantitative and qualitative results unequivocally demonstrate that our approach surpasses previous methods and heuristic baseline by a significant margin.
\noindent
To summarize, our main contributions are as follows:
\begin{itemize}
  \item A novel formulation to quantify the observed information and Information Gain for Radiance Fields.
  \item An effective and efficient view selection method exploiting the sparse structure of the scene rendering problem.
  \item An efficient pixel-wise uncertainty quantification and visualization method for 3D Gaussian Splatting.
  \item Extensive comparative study showing that our method outperforms existing active approaches on view selection, active mapping, and uncertainty quantification. 
\end{itemize}

\section{Related Works}
\label{sec:related_works}

\noindent
Active Learning and Radiance Fields are prosperous research fields with various research directions. In this section, we limit our literature review to works that directly relate to our tasks. We refer the readers to literature reviews if they are interested in Radiance Fields~\cite{gao2023nerf-review, dellaert2021neural-review} or Deep Active Learning~\cite{ren2021survey,zhan2022comparative}.
Besides, Fisher Information has been extensively studied in deep active learning~\cite{Ash2021GoneFishing,Ash2020Deep,kothawade2021similar,kirsch2021stochastic}.
Notably, Kirsch~\etal~\cite{kirsch2022unifying} unified existing works in active learning for deep learning problems from the perspective of Fisher Information, which shares many insights with our method. 
\paragraph{Uncertainty Quantifications for Radiance Fields}
In the Neural Radiance Fields, Lee~\etal~\cite{lee2022uncertainty}, Yan~\etal~\cite{yan2023activeIO} and Zhan~\etal~\cite{zhan2022activermap} attempted to quantify the uncertainty in a scene by the distribution of densities on a casted ray. Shen~\etal~\cite{CF-NeRF,shen2021snerf} designed Bayesian models by re-parametrizing the NeRF model. However, they only tackled the predictive uncertainty of the model that did not relate to the observed information of the parameters. 
Sunderhauf~\etal\cite{sünderhauf2022densityaware} propose an additional uncertainty measure in uncharted regions, determined by the ray termination probability on the learned geometry.
Regarding concurrent work, Goli~\etal~\cite{goli2023} introduces a hypothetical perturbation field and interpolates uncertainty over NeRF-style query points for uncertainty quantification and NeRF clean-up task.
The perturbation field with respect to query points is also not compatible with recent radiance field models like 3D Gaussian Splatting~\cite{kerbl3Dgaussians}, which do not take query points as inputs.
Besides, their algorithm is not designed for an active learning problem since selecting candidates with the largest per-pixel uncertainty may not be optimal, as the information gain for the model parameters is not concerned.
In contrast, we focus on active learning problems. We derive our theory from the expected information gain for the model and compute the Hessian for the model parameters, allowing us to apply our algorithm to recent advancements in radiance fields such as 3D Gaussian splatting. 
Our paper focuses on active view selection and mapping based on our novel expected information gain objective. 
Moreover, their implementation relies on multiple function calls to PyTorch backward engine whereas our efficient CUDA implementation for the diagonal Hessian computation consumes much less time (11.3 ms $\pm$ 33.3 $\mu$s v.s. 1.1 s $\pm$ 2.5 ms).
To the best of our knowledge, we are the first method that quantifies the uncertainty directly on model parameters for Radience Fields, thanks to the powerful framework of 3D Gaussian and our efficient formulation. 
\paragraph{Active View Selection and Mapping}
Although the next best view selection problem was extensively studied before radiance fields gained popularity, there has not been much literature on active ``training'' view selection for volumetric rendering. 
Jin~\etal~\cite{jin2023neu} trains a neural network to predict per-pixel uncertainty for view selection. However, their method requires an image capture as the input of the network, which is closer to dataset subsampling rather than active learning.
ActiveNeRF~\cite{pan2022activenerf} studied the next best view selection by directly adding variances as the output of the vanilla NeRF, and it is the closest approach to the view selection problem we are working on. 
In a larger scope, Active Perception~\cite{bajcsy1988active,bajcsy2018revisiting} has been studied along with the advancement of robotics and computer vision.
Traditionally, frontier-based approach~\cite{yamauchi1997frontier} and Rapidly exploring Random Tree~(RRT)~\cite{lavalle1998rrt} are the most representative and have been extended and enhanced for different data representations~\cite{zhu20153d, karaman2010incremental, dornhege2013frontier, shen2012autonomous} and complex conditions~\cite{ye2022multi, papachristos2017uncertainty}.
For the larger scope of Active Simultaneous Localization and Mapping~(SLAM), please refer to literature reviews~\cite{lluvia2021active-survey, placed2023survey}.
In recent years, Guedon~\etal~\cite{guedon2022scone,guedon2023macarons} attempted to improve the coverage for mapping systems using point cloud representation with neural networks. Dhami~\etal~\cite{dhami2023pred} studied the next best view selection in view planning by first completing the point cloud from partial observation. Georgakis~\etal~\cite{georgakis2022uncertainty} measured uncertainty through deep ensemble. Chaplot~\etal~\cite{chaplot2020learning} and Ramakrishnan~\etal~\cite{ramakrishnan2020occupancy} developed active exploration methods with deep reinforcement learning.
The mapping system that uses implicit feild~\cite{zhu2021niceslam, Sucar:etal:ICCV2021, Ortiz:etal:iSDF2022, sandstrom2023point} espeically 3D Gaussians~\cite{keetha2023splatam, yan2023gs, Matsuki:Murai:etal:CVPR2024,yugay2023gaussianslam} as representation has been widely studied.
This presents the need for active mapping algorithms that could maximize the mapping efficiency in an embodied robot system.
Similar to ActiveNeRF~\cite{pan2022activenerf}, Ran~\etal~\cite{Ran2023neurar} performed trajectory planning for object-centric autonomous reconstruction systems by predicting variance as an extra output of the network. 
Yan~\etal\cite{yan2023active-neural-mapping} discussed the intuition of flat minimum and quantified the predictive uncertainty of neural mapping models through the lens of loss landscape. However, they only approximate the uncertainty of the variances from the output of neighboring points. Our method directly quantifies the observed information of the parameters by computing the Hessian matrix of the log-likelihood function of the model's objective.
Besides, we introduce the Expected Information Gain for mapping systems from the first principle as a complementary method to previous heuristics. 

\section{Technical Approach} \label{sec:method}

In this section, we first introduce how we use Fisher Information in Sec.~\ref{sec:fisher-render}. By leveraging Fisher Information, we demonstrate how to select the next best view and next best batch of views in Sec.~\ref{sec:nbvs} and Sec.~\ref{sec:batch}. We then extend our method to large-scale scenes with active mapping. Finally, we showcase that our method could quantify the pixel-wise uncertainties in Sec.~\ref{sec:pixel-uncern}.
\subsection{Fisher Information in Volumetric Rendering}\label{sec:fisher-render}
Fisher Information is a measurement of information that an observation $(\bx, \by)$ carries about the unknown parameters $\bw$ that model $p(\by|\bx; \bw)$. In the problem of novel view synthesis,  $(\bx, \by)$ are the camera pose $\bx$ and image observation $\by$ at pose $\bx$, respectively, whereas $\bw$ are the parameters of the radiance field.
The objective of neural rendering is equivalent to minimizing the negative log likelihood~(NLL) between rendered images and ground truth images on the holdout set, which inherently represents the quality of scene reconstruction
\begin{equation}\label{logp}
   - \log p(\by|\bx, \bw) = \left(\by - f(\bx, \bw)\right)^T \left(\by - f(\bx, \bw)\right)
\end{equation}
where $f(\bx, \bw)$ is our rendering model. 
Under regularity conditions~\cite{schervish2012theory}, the Fisher Information of the model $\log p(\by|\bx; \bw)$ is the Hessian of the log-likelihood function with respect to the model parameters $\bw$:
\begin{equation}
     \mathcal{I}(\bw) = - \mathbb{E}_{\log p(\by|\bx, \bw)} \big[\frac{\partial^2 \log p(\by|\bx, \bw)}{\partial \bw^2} \big| \bw \big] = \bH''[\by|\bx, \bw]
\end{equation}
where $\bH''[\by|\bx, \bw]$ is the Hessian matrix of Eq.~(\ref{logp}).

\subsection{Next Best View Selection Using Fisher Information}\label{sec:nbvs}
In the active view selection problem, we start with a training set $D^{train}$ and have an initial estimation of parameters $\bw^*$ using $D^{train}$. The aim is to select the next best view that maximizes the Information Gain~\cite{lindleyEIG, HoulsbyBAL,KirschBatchBALD} between candidates views $\bxa \in D^{pool}$ and $D^{train}$, where $D^{pool}$ is the pool of candidate views:
\begin{align}
    & \mathcal{I}[\bw^*; \{\bya\}|\{\bxa\}] \nonumber \\
    & = \mathbb{H}[\bw^* | D^{train}] - \mathbb{H} [\bw^*| \{\bya\}, \{\bxa\}, D^{train}] 
    \label{eq:entropy-acq}
\end{align}
where $H[\cdot]$ is the entropy~\cite{kirsch2022unifying}.

When the log-likelihood has the form of Eq.~(\ref{logp}), in our case, the rendering error, the difference of the entropies in the R.H.S. of Eq.~(\ref{eq:entropy-acq}) can be approximated as~\cite{kirsch2022unifying}: 
\begin{align}\label{eq:entropy-log}
 \frac{1}{2} \log \det \left(\bH''[\{\bya\}|\{\bxa\}, \bw^*]\; \bH''[\bw^*| D^{train}]^{-1} + I \right) \nonumber\\
    \leq \frac{1}{2}  \tr\left( \bH''[\{\bya\}|\{\bxa\}, \bw^*]\; \bH''[\bw^*| D^{train}]^{-1}\right).
\end{align}
As Fisher Information is additive, $\bH''[\bw^*| D^{train}]^{-1}$ can be computed by summing the Hessians of model parameters across all different views in $\{D_{train}\}$ before inverting. We can choose the next best view $\bxa$ by optimizing
\begin{equation}\label{eq:opt}
    \argmax_{\bxa} \tr\left( \bH''[\bya | \bxa, \bw^*] \; \bH''[\bw^*| D^{train}]^{-1}\right).
\end{equation}
The Hessian $\bH''[\by|\bx, \bw^*]$ of our model can be computed as:
\begin{equation}\label{eq:hessian-full}
    \bH''[\by|\bx, \bw^*] = \nabla_\bw f(\bx; \bw^*) ^T \nabla_{f(\bx;\bw^*)}^2 H[\by| f(\bx;\bw^*)] \nabla_\bw f(\bx; \bw^*)
\end{equation}
where $\bH''[\by| \bx,\bw^*]$ in our case is equal to the covariance of the RGB measurement that we set equal to one.
Hence, the Hessian matrix can be computed just from the Jacobian matrix of $f(\bx, \bw)$
\begin{equation}\label{eq:hessian-simpilifed}
    \bH''[\by|\bx, \bw^*] = \nabla_\bw f(\bx; \bw^*) ^T \nabla_\bw f(\bx; \bw^*).
\end{equation}
We can optimize the objective in Eq.~(\ref{eq:opt}) without knowing the ground truth of candidate views $\{\bya\}$, which was expected since the Fisher Information never depends on the observations themselves.
The Hessian in~(\ref{eq:hessian-simpilifed})
has only a limited number of off-diagonal elements because each pixel is considered independent in $- \log p(\by|\bx, \bw)$.
Furthermore, recent NeRF models~\cite{plenoxels,mueller2022instant, reiser2021kilonerf,sun2021direct} typically employ structured local parameters that each parameter only contributes to the radiance and density of a limited spatial region for faster convergence and rendering. Therefore, only parameters that contribute to the color of the pixels would share non-zero values in the Hessian matrix $ \bH''[\by|\bx, \bw^*]$. 
However, the number of optimizable parameters is typically more than 200 million, which means it is impossible to compute without sparsification or approximation. 
In practice, we apply Laplace approximation~\cite{laplace2021, bayesian-interpolation} that approximates the Hessian matrix with its diagonal values plus a log-prior regularizer $\lambda I$
\begin{equation}\label{eq:hessian-reg}
    \bH''[\by|\bx, \bw^*] \simeq \diag(\nabla_\bw f(\bx, \bw^*) ^T \nabla_\bw f(\bx, \bw^*)) + \lambda I.
\end{equation}
\subsection{Batch Active View Selection}
\label{sec:batch}
Selecting multiple views to capture new images is useful for its possible applications, such as view planning and scene reconstruction. 
If we simply use Eq.~(\ref{eq:opt}) to select a batch of acquisition samples $\{\bxa\}$, we could possibly select very similar views inside the acquisition set $\{\bxa\}$ as we do not consider the mutual information between our selections. 
However, we would face a combinatorial explosion if we directly attempt to maximize the expected information gain between training and acquisition samples and simultaneously minimize the mutual information across acquisition samples. Therefore, we employ a greedy optimization algorithm as illustrated in Alg.~\ref{alg:batch}, which is $\nicefrac{1}{e}$-approximate in Fisher Information~\cite{nemhauser1978analysis}. When batch size $B$ is 1, our algorithm is equivalent to Eq.~(\ref{eq:opt}). 
Please note that we focus on batch active view selection instead of dataset subsampling as it is more related to real-world scenarios where we wish to plan a trajectory for an agent to acquire more training views. 

\begin{algorithm}\caption{Batch Active Views Selection}\label{alg:batch}
	\LinesNumbered
	\KwIn{$\{\bH''[\bw^*|\bxa]\}$, $\bH''[\bw^*| D^{train}]$, number of views to select $B$}
	\KwOut{Selected Views $S_B$}

        $S_0 \leftarrow \emptyset$ \;
        $H_0 \leftarrow \bH''[\bw^*| D^{train}]$\;

	\For { $b \leftarrow 1$ \KwTo $B$}{
 	$i \leftarrow \argmax_{ \bxa \in  D^{pool}] \setminus S_{b-1} } \tr\left( \bH''[\bya | \bxa, \bw^*] \; H_{b-1}^{-1}\right)$\;
            $S_{b} \leftarrow S_{b-1} \cup \{ i \}$   \;
            $H_{b} \leftarrow H_{b-1} +  \bH''[\bya | \bxa, \bw^*] $  \;
        }
\end{algorithm}
\subsection{Active Mapping with 3D Gaussian Splatting}
\label{sec:mapping}
\begin{figure}
    \centering
\includegraphics[width=\textwidth]{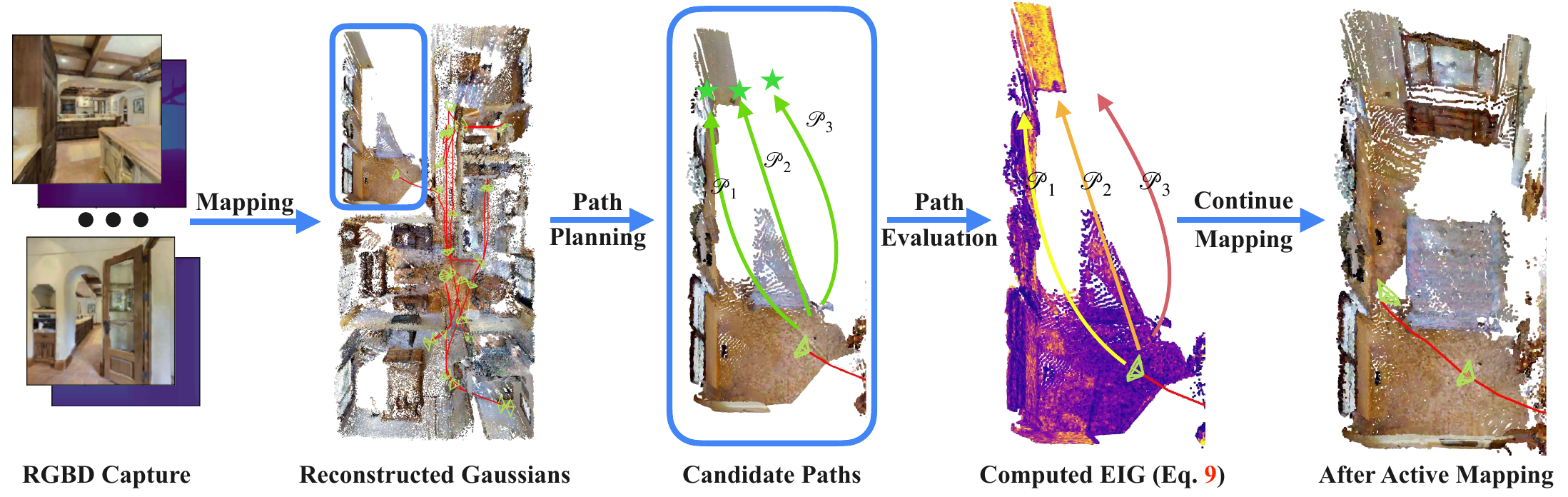}
    \vspace{-5mm}
    \caption{{\bf An Illustration of Our Active Mapping System} Given RGBD captures, we first reconstruct the environment using 3D Gaussians as representation. Afterward, we select a set of goal points from map frontiers and plan the shortest path to each goal. The EIG is computed for each path, and we choose the one with the highest EIG to continue exploration.   }
    \label{fig:pipeline}
    \vspace{-5mm}
\end{figure}

With the batch selection algorithm discussed in Sec.~\ref{sec:batch}, our method can be extended as an active mapping algorithm. By maximizing the EIG from candidate trajectories $\{\mathcal{P}_j\}$, we can determine the most informative path for exploration: 
\begin{equation}\label{eq:path}
    \argmax_{\mathcal{P}_j} \tr\left( \left( \sum_{\bxa \in \mathcal{P}_j} \bH''[\bya | \bxa, \bw^*]\right) \; \bH''[\bw^*| D^{train}]^{-1}\right).
\end{equation}
We build our active mapping system on top of existing passive mapping architecture SplaTAM~\cite{keetha2023splatam} that uses 3D Gaussians as basic data representation.
We employ frontier-based exploration~\cite{yamauchi1997frontier} to propose path candidates and sample cameras along the path concerning the mobility of the agent. 
The pipeline is illustrated in Fig. \ref{fig:pipeline}.
Our path selection algorithm is compatible with other active mapping systems and can be used as an extension to select the most informative path. 
\subsection{Pixel-wise Uncertainty with Volumetric Rendering}\label{sec:pixel-uncern}
Previously, we discussed quantifying the Information Gain for any camera views. Our model can also be adapted to obtain pixel-wise uncertainties.
As we have discussed in Sec.~\ref{sec:nbvs}, we can approximate the uncertainty on each parameter with the diagonal elements of the hessian matrix $\bH''[\bya | \bxa, \bw^*]$. In recent Radiance Field models~\cite{kerbl3Dgaussians, plenoxels}, the parameters directly correspond to a spatial location in the scene.
Therefore, we can compute the uncertainty in our rendered pixels by examining the diagonal Hessians along the casted ray for volumetric rendering
\begin{equation}
\label{eq:render-uncern}
\mathbf{U}(\text{r}) = \sum_{n=1}^{N_s} T_i \left(1-\exp(-\sigma_n \delta_n) \right) tr({\bf G}_n)
\end{equation}
where ${\bf G}_n$ is the submatrix of $\bH''[\bw^*| D^{train}]$ containing the rows and columns that correspond to parameters at location $n$.
Please note that this is a relative uncertainty derived from the observed information on the model parameters, which does not involve metric scales. If we wish to estimate absolute uncertainties on predictions like depth maps, we need to denormalize the uncertainty in each term by its depth $d_n$.

\section{Experiments} \label{sec:exps}
In this Section, we present the empirical evaluation of our approach.
The running time analysis for our customized CUDA kernel and other implementation details can be found in the supplementary.
We first focus on the experiments of active view selection and compare our method quantitatively and qualitatively against the previous method (Subsection~\ref{sec:view-exp}).
We further present our results on active mapping (Subsection~\ref{sec:mapping-exp}).
Finally, we present more results of our method on uncertainty quantification results (Subsection~\ref{sec:exp_uncern}).

\subsection{Active View Selection}\label{sec:view-exp}
We conducted extensive experiments on view selections to demonstrate that our expected information gain could help the model find the next best views. Here, we first introduce the dataset we use and detailed experimental settings. Then, we compare our method with random baselines and previous state-of-the-art ActiveNeRF~\cite{pan2022activenerf} quantitatively and qualitatively.
\paragraph{Datasets} Our approach is extensively evaluated on two common benchmark datasets: Blender Dataset~\cite{mildenhall2020nerf} and the challenging real-world Mip-NeRF360 dataset~\cite{barron2022mipnerf360}. The Blender dataset comprises eight synthetic objects with intricate geometry and realistic non-Lambertian materials. Each scene in this dataset includes 100 training and 200 test views, all with a resolution of $800\times800$. Our method uses the 100 training views as a candidate pool to select training views, and we evaluate all the models on the full 200-view test set.
We use the default training configuration as in 3D Gaussian and Plenoxels for this dataset.
Mip-NeRF360~\cite{barron2022mipnerf360} is a real-world $360^\circ$ dataset captured for nine different scenes. It has been widely used as a quality benchmark for novel view synthesis models~\cite{barron2023zipnerf, multinerf2022}. We train our models at the resolution of $1066\times1600$ following 3D Gaussian Splatting~\cite{kerbl3Dgaussians}.
\paragraph{Metrics} Our evaluations utilize image quality metrics such as peak signal-to-noise ratio~(PSNR) and structural similarity index~(SSIM)~\cite{wang2003multiscale}. Additionally, we incorporate LPIPS \cite{zhang2018unreasonable}, which provides a more accurate reflection of human perception.
\paragraph{Baselines} 
We quantitatively and qualitatively compare our method against the current state-the-art ActiveNeRF~\cite{pan2022activenerf} and random selection baseline.
To make a fair comparison with  ActiveNeRF~\cite{pan2022activenerf},
We re-implemented a similar variance estimation algorithm in 3D Gaussian Splatting and Plenoxels in CUDA. We assign each 3D Gaussian (or grid cell in Plenoxels) a variance parameter and use volumetric rendering to render a variance map. The supplementary materials provide more details about our re-implementation. 
To show the difference between EIG and uncertainty, we adapt the uncertainty estimation method BayesRays~\cite{goli2023} into the view selection algorithm by directly using the rendered uncertainty value on each candidate view as a metric for view selection. 
 
\begin{figure*}[!t]
\setlength{\tabcolsep}{0pt}
	\centering
\begin{tabular}{p{0.3cm}ccccc}
\rotatebox[origin=c]{90}{\parbox{\widthof{AcNeRF\ }}{ACNeRF}} & \raisebox{-0.5\height} {\includegraphics[width=0.19\linewidth]{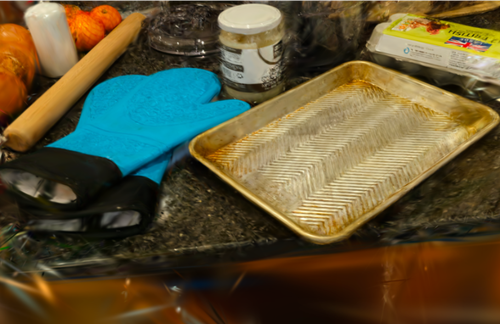}} & \raisebox{-0.5\height} {\includegraphics[width=0.19\linewidth]{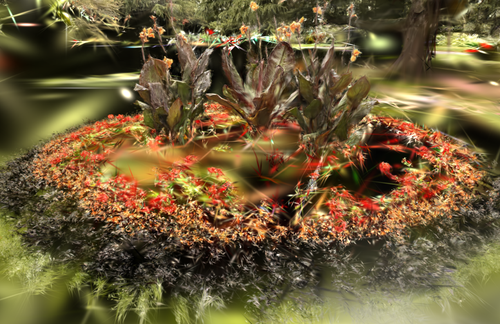}}  & \raisebox{-0.5\height} {\includegraphics[width=0.19\linewidth]{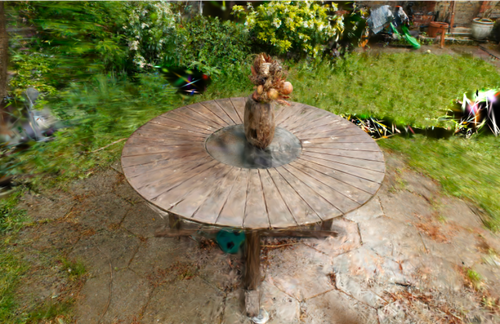}}  & \raisebox{-0.5\height} {\includegraphics[width=0.19\linewidth]{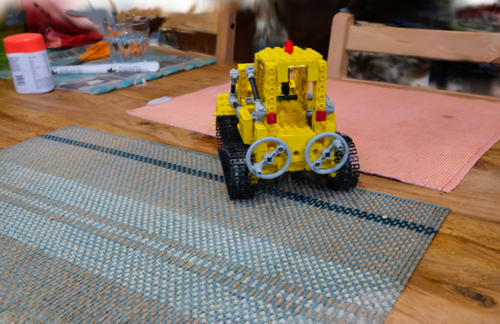}}  & \raisebox{-0.5\height} {\includegraphics[width=0.19\linewidth]{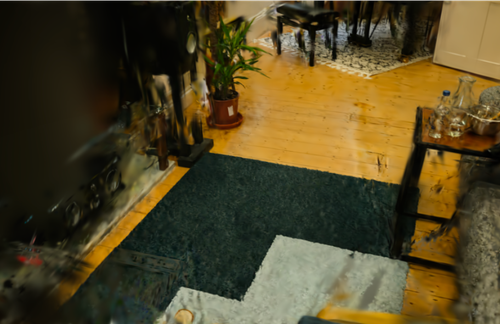}} \\
\rotatebox[origin=c]{90}{Random} & \raisebox{-0.5\height} {\includegraphics[width=0.19\linewidth]{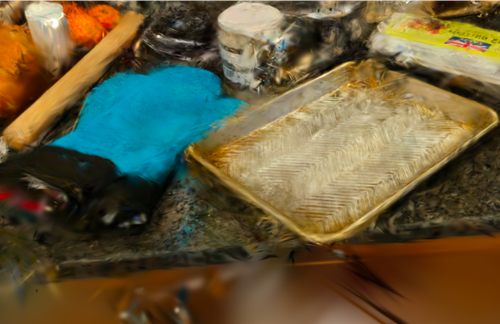}} & \raisebox{-0.5\height} {\includegraphics[width=0.19\linewidth]{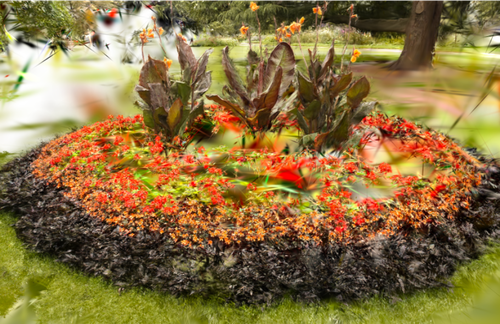}}  & \raisebox{-0.5\height} {\includegraphics[width=0.19\linewidth]{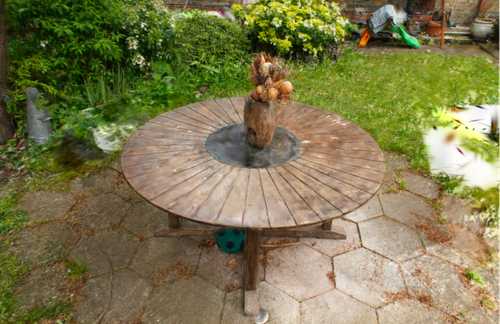}}  & \raisebox{-0.5\height} {\includegraphics[width=0.19\linewidth]{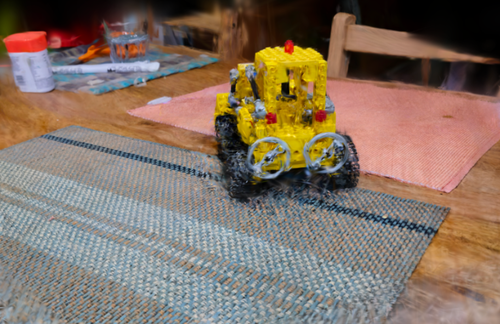}}  & \raisebox{-0.5\height} {\includegraphics[width=0.19\linewidth]{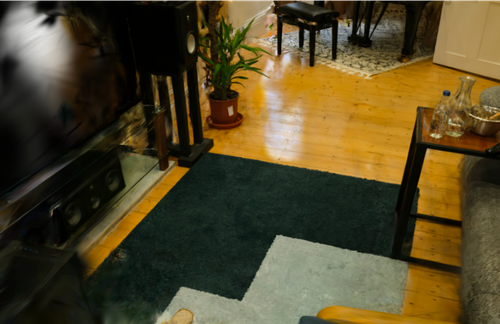}} \\
\rotatebox[origin=c]{90}{Ours} & \raisebox{-0.5\height} {\includegraphics[width=0.19\linewidth]{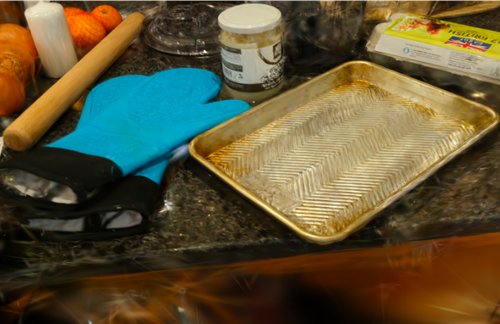}} & \raisebox{-0.5\height} {\includegraphics[width=0.19\linewidth]{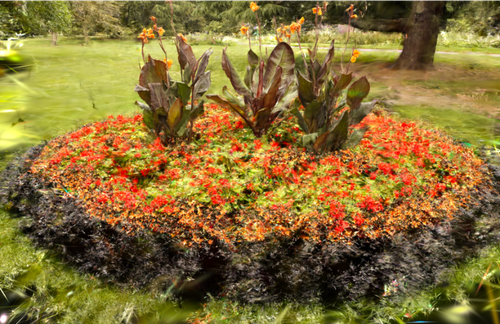}}  & \raisebox{-0.5\height} {\includegraphics[width=0.19\linewidth]{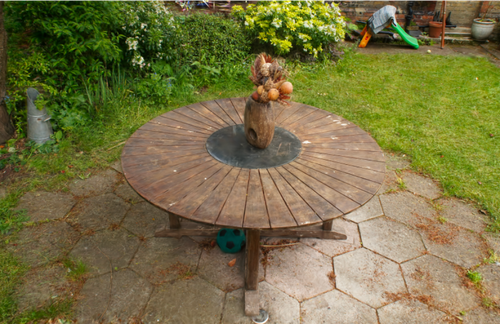}}  & \raisebox{-0.5\height} {\includegraphics[width=0.19\linewidth]{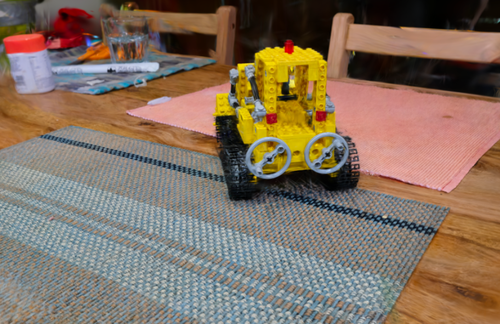}}  & \raisebox{-0.5\height} {\includegraphics[width=0.19\linewidth]{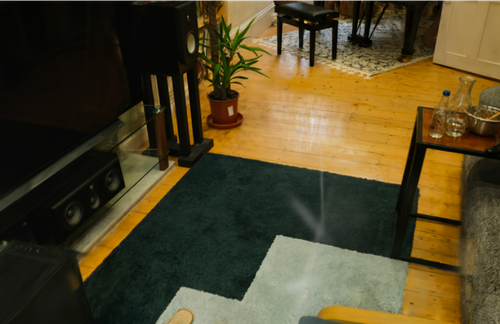}} \\
\rotatebox[origin=c]{90}{GT} & \raisebox{-0.5\height}{\includegraphics[width=0.19\linewidth]{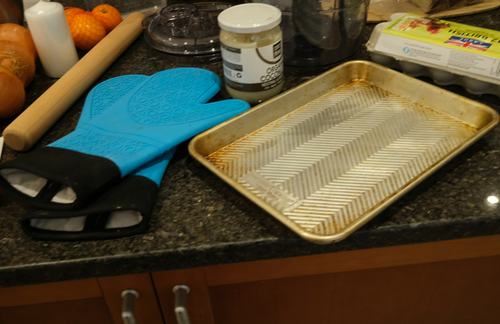}} & \raisebox{-0.5\height} {\includegraphics[width=0.19\linewidth]{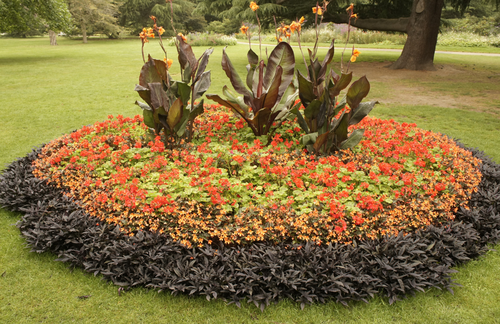}}  & \raisebox{-0.5\height} {\includegraphics[width=0.19\linewidth]{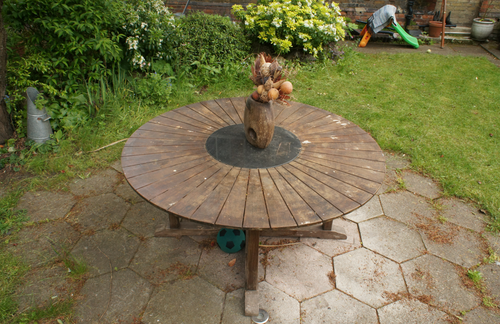}}  & \raisebox{-0.5\height} {\includegraphics[width=0.19\linewidth]{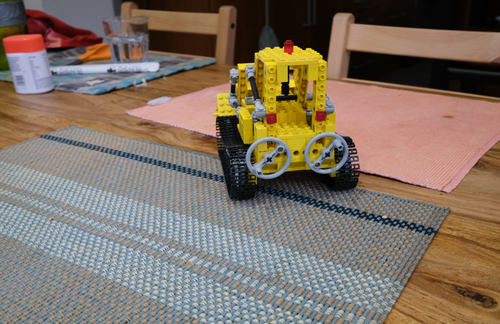}}  & \raisebox{-0.5\height} {\includegraphics[width=0.19\linewidth]{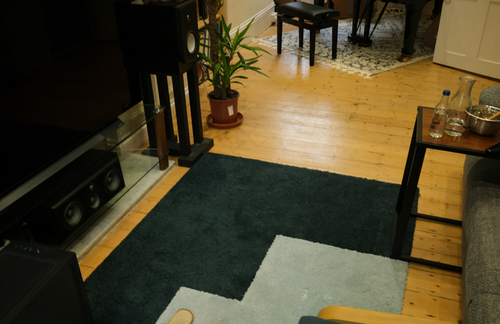}} \\
\end{tabular}
	\vspace{-2mm}
	\caption{{\bf Qualitative Study of our method on Mip360 Dataset} From the top to bottom are results from ActiveNeRF, random baseline, our method, and the ground truth. All the models in this figure are implemented on top of 3D Gaussian Splatting~\cite{kerbl3Dgaussians} for better performance on this challenging dataset. We could see baseline models exhibited artifacts in some renderings due to their lack of constraints from nearby training views.  
}
\label{fig:mip}
\vspace{-5mm}
\end{figure*}

\begin{figure}[t]
\centering
\begin{tabular}{cccccccc}
\multicolumn{4}{c}{\textbf{20 Training Views}} & \multicolumn{4}{c}{\textbf{10 Training Views}}\\
ActiveNeRF & Random & BayesRays & Ours & ActiveNeRF & Random & BayesRays & Ours \\
\includegraphics[width=0.12\linewidth]{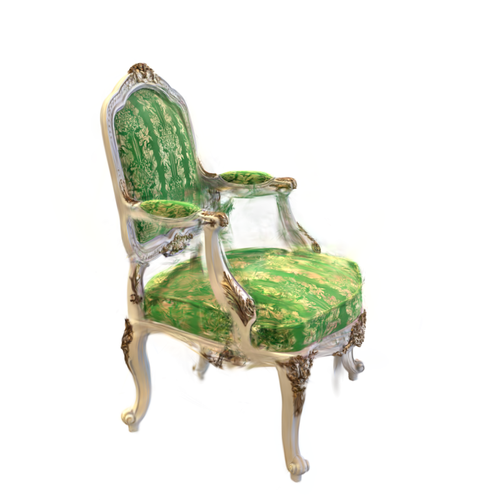} & \includegraphics[width=0.12\linewidth]{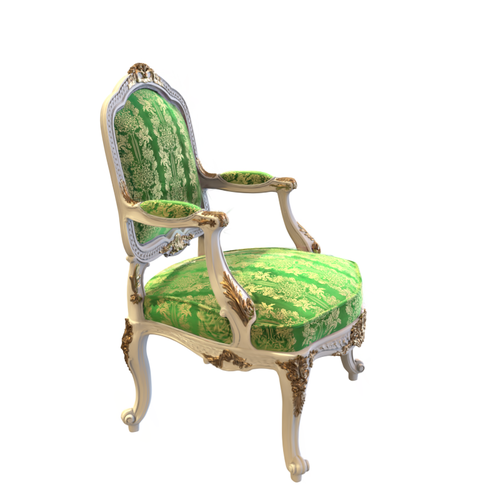} & \includegraphics[width=0.12\linewidth]{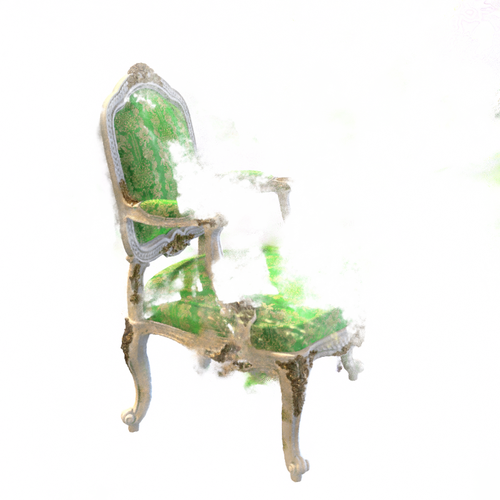} & \includegraphics[width=0.12\linewidth]{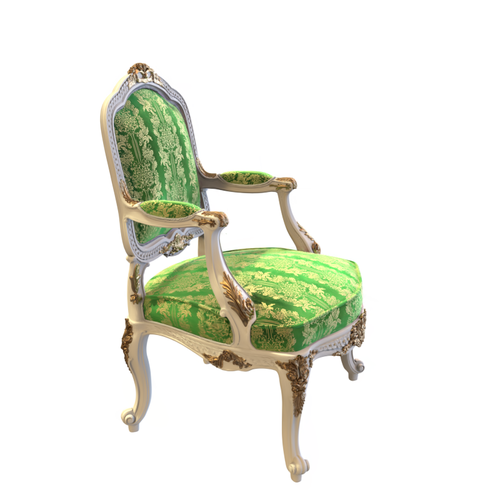} &\includegraphics[width=0.12\linewidth]{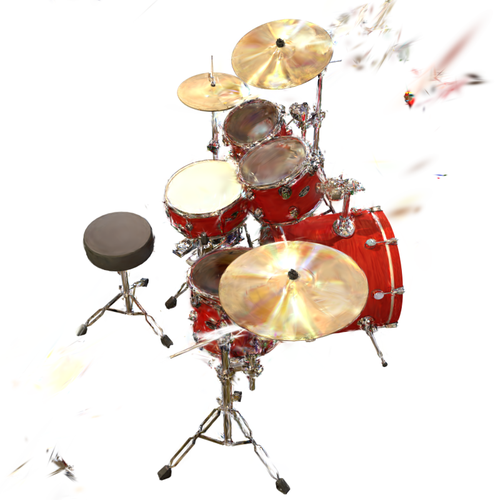} & \includegraphics[width=0.12\linewidth]{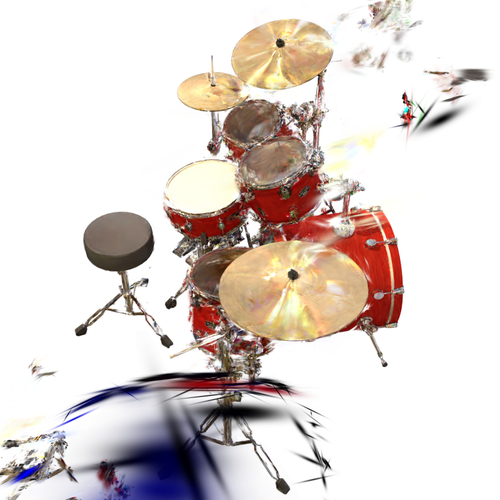} & \includegraphics[width=0.12\linewidth]{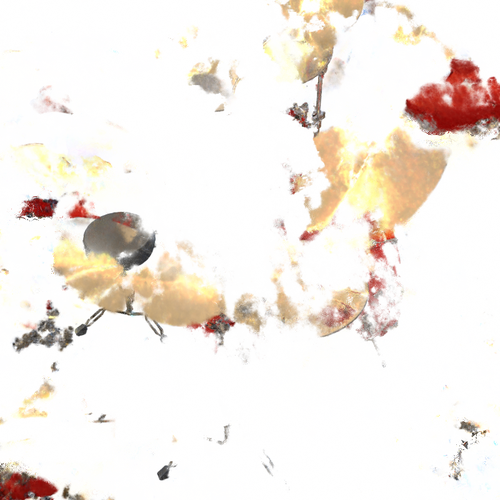} & \includegraphics[width=0.12\linewidth]{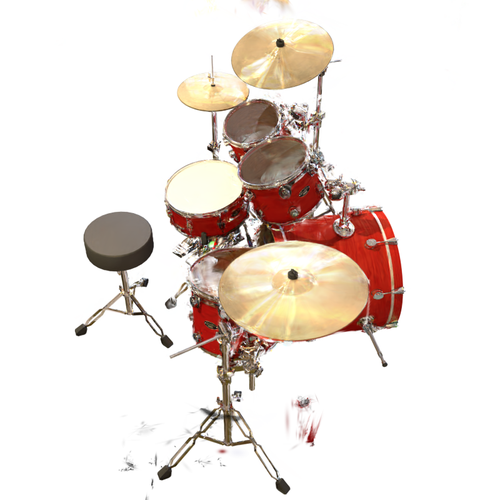}\\
\includegraphics[width=0.12\linewidth]{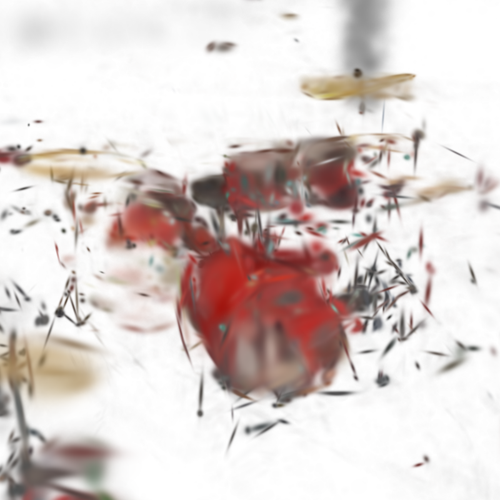} & \includegraphics[width=0.12\linewidth]{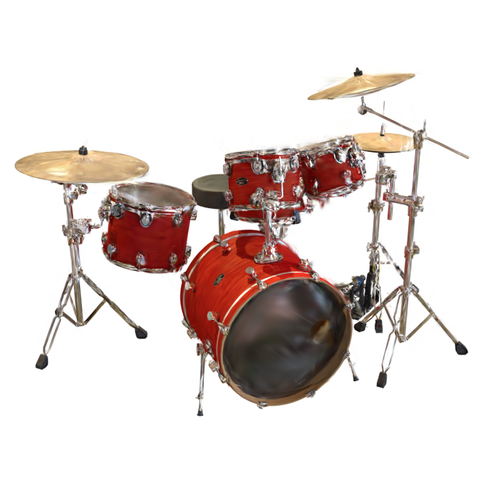} & \includegraphics[width=0.12\linewidth]{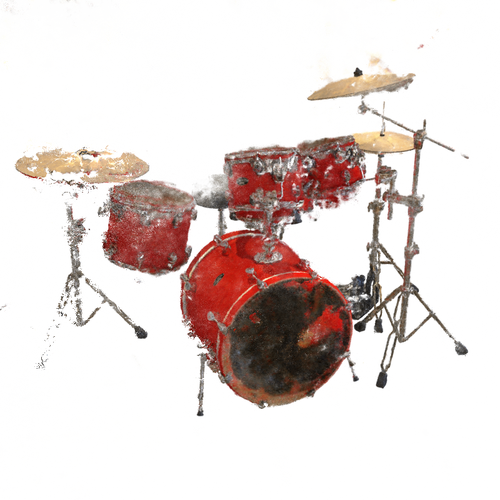} & \includegraphics[width=0.12\linewidth]{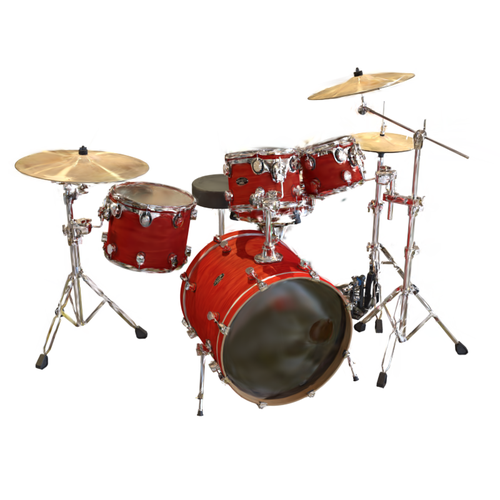} & \includegraphics[width=0.12\linewidth]{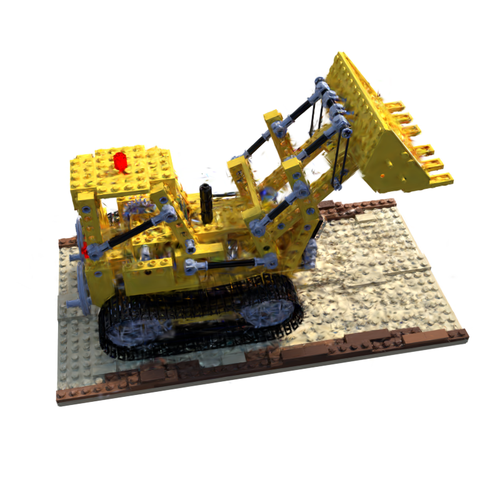} & \includegraphics[width=0.12\linewidth]{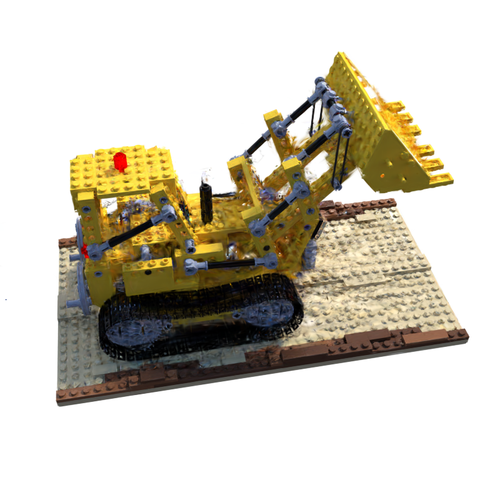} & \includegraphics[width=0.12\linewidth]{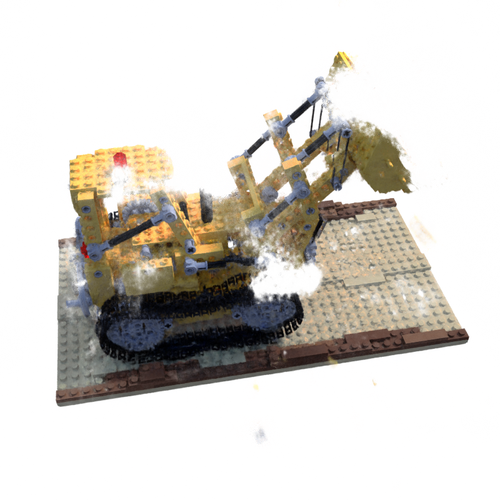} & \includegraphics[width=0.12\linewidth]{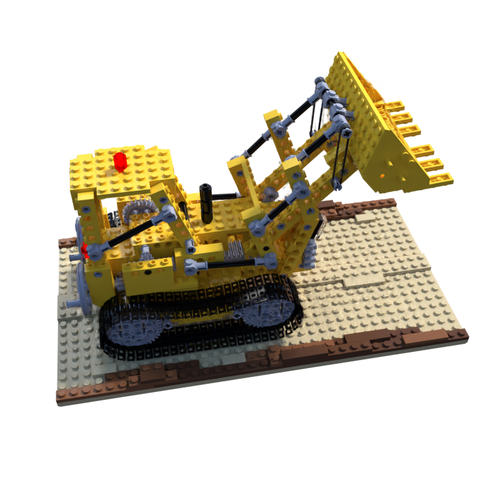}\\
\includegraphics[width=0.12\linewidth]{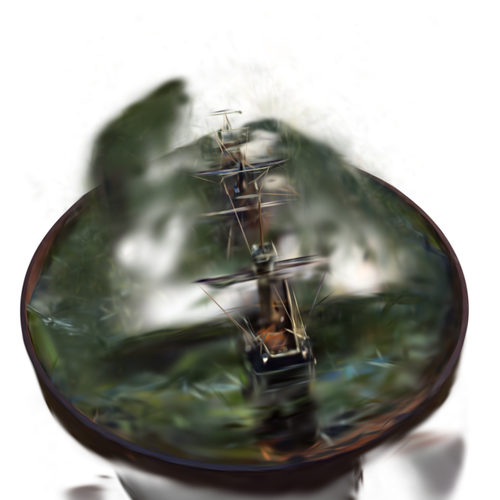} & \includegraphics[width=0.12\linewidth]{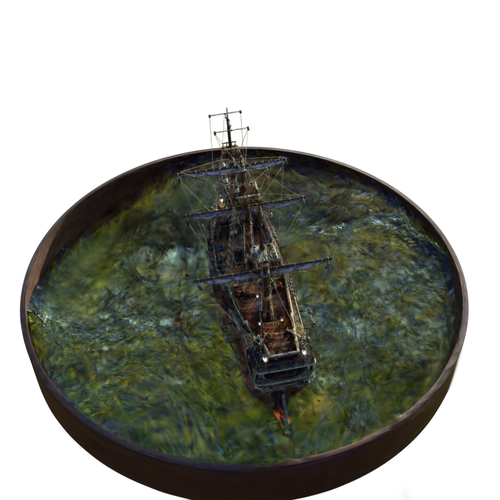} & \includegraphics[width=0.12\linewidth]{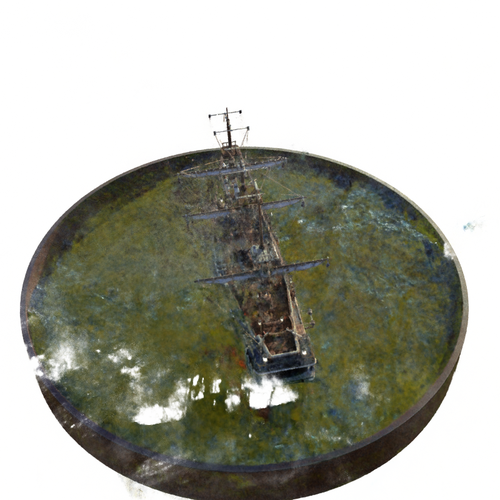} & \includegraphics[width=0.12\linewidth]{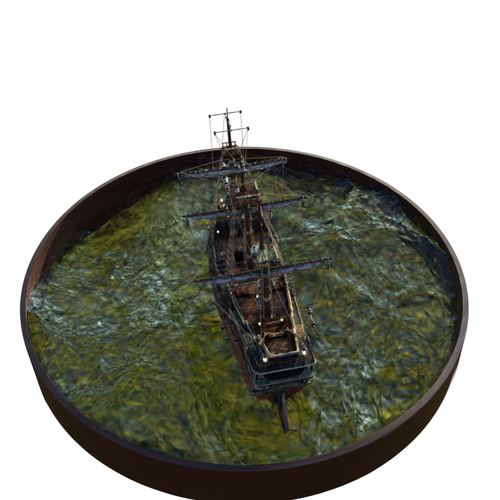} & \includegraphics[width=0.12\linewidth]{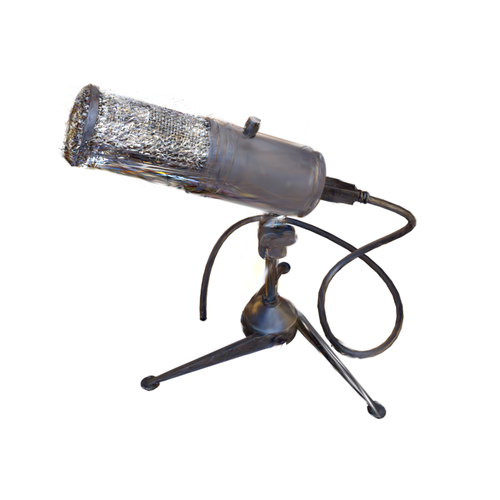} & \includegraphics[width=0.12\linewidth]{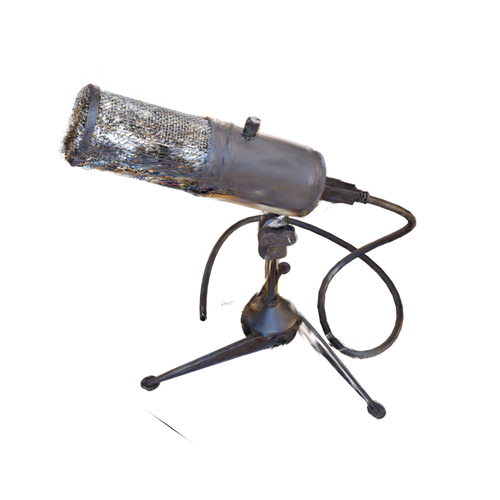} & \includegraphics[width=0.12\linewidth]{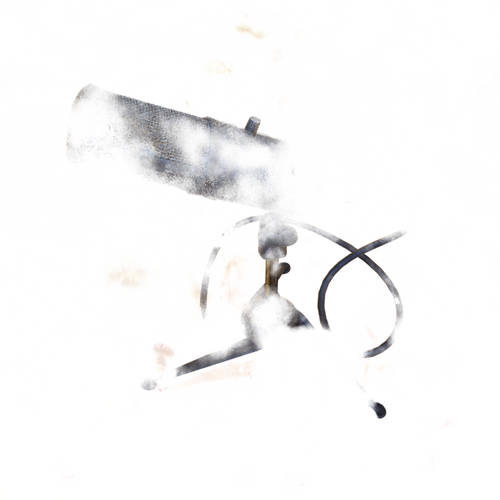} & \includegraphics[width=0.12\linewidth]{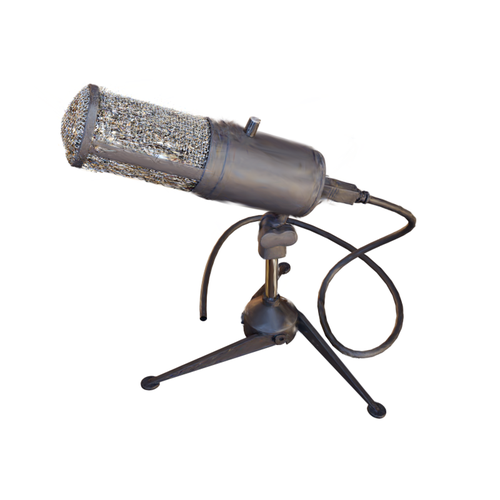}\\
\end{tabular}
\vspace{-3mm}
\caption{{\bf Qualitative Results on Blender Dataset with 20 and 10 Training Views.} 
All the methods are implemented on 3D Gaussian Splatting and compared in the same training configuration except for different training views selected by different methods.
}\label{fig:qual-blender}
\vspace{-5mm}
\end{figure}

\paragraph{Experiment Settings} We experiment with our model with 3D Gaussian Splatting backend on both the next view selection and the next batch of view selections across both the Blender and Mip360 Dataset. Each model is initialized with the same random seed and was trained on the same four uniform views. Each model is trained for 30,000 iterations following the default configurations of 3D Gaussian Splatting~\cite{kerbl3Dgaussians}. Similar to the training program of Gaussian Splatting, we reset the opacity every time we select new views to avoid degeneration of the training procedure. 
All the external settings in the experiment are kept the same except for the view selection algorithms.

Similarly, we also showcase the implementations of our active view selection algorithm on Plenoxels in the Blender Dataset. The experimental settings for initial views and view selection schedules are the same, except view selection was made every four epochs.
\noindent
\begin{itemize}
    \item Sequential Active View Selection: 1 new view is selected every 100 epochs till the model has 20 training views. 
    \item Batch Active View Selection: 4 new views are selected every 300 epochs till the model has 20 training views.
\end{itemize}
\definecolor{tabfirst}{rgb}{1, 0.7, 0.7} %
\definecolor{tabsecond}{rgb}{1, 0.85, 0.7} %
\definecolor{tabthird}{rgb}{1, 1, 0.7} %
\begin{table}[t]
\begin{minipage}{0.47\textwidth}
    \centering
    \caption{{\bf Active View Selections on Blender Dataset} The best, second, and third results are highlighted in red, orange, and yellow, respectively.\;  $*$: Numbers are taken directly from ActiveNeRF~\cite{pan2022activenerf} as reference. }
    \label{tab:blender20}
    \resizebox{\textwidth}{!}{
    \begin{tabular}{l|ccc}
    Method & PSNR	 $\uparrow$ &	SSIM  $\uparrow$	&	LPIPS $\downarrow$\\
    \hline
    
ActiveNeRF*              &                      26.240 &                      0.856 &                      0.124  \\
Plenoxels + Random        &                      23.906 &                      0.863 &                      0.174  \\
Plenoxels + ActiveNeRF    &                      23.522 &                      0.857 &                      0.150  \\
Nerfacto + Random        &                      22.614 &                      0.890 &                      0.111  \\
Nerfacto + BayesRays     &                      22.504 &                      0.887 &                      0.115  \\
3D Gaussian + Random     & \cellcolor{tabsecond}28.732 & \cellcolor{tabsecond}0.939 & \cellcolor{tabsecond}0.053  \\
3D Gaussian + ActiveNeRF &  \cellcolor{tabthird}26.610 &  \cellcolor{tabthird}0.905 &  \cellcolor{tabthird}0.081  \\
\hline
Plenoxel + Ours          &                      24.513 &                      0.876 &                      0.157  \\
3D Gaussian + Ours       &  \cellcolor{tabfirst}29.525 &  \cellcolor{tabfirst}0.944 &  \cellcolor{tabfirst}0.043 

    \end{tabular}
    }
\end{minipage}
    ~
\begin{minipage}{0.47 \textwidth}
    \centering
    \caption{{\bf Active Batch View Selections on Blender Dataset} Each time, 4 views are selected based on different view selection methods. $*$: Numbers are taken directly from ActiveNeRF~\cite{pan2022activenerf} as reference. }
    \label{tab:blender-batch}
    \resizebox{\textwidth}{!}{
    \begin{tabular}{l|ccc}
    Method & PSNR	 $\uparrow$ &	SSIM  $\uparrow$	&	LPIPS $\downarrow$\\
    \hline
    
ActiveNeRF*              &                      26.240 &                      0.856 &                      0.124  \\
Plenoxels + Random        &                      23.242 &                      0.862 &                      0.158  \\
Plenoxels + ActiveNeRF    &                      23.147 &                      0.857 &                      0.148  \\
Nerfacto + Random        &                      21.622 &                      0.875 &                      0.137  \\
Nerfacto + BayesRays     &                      22.160 &                      0.878 &                      0.127  \\
3D Gaussian + Random     &  \cellcolor{tabthird}27.135 & \cellcolor{tabsecond}0.927 & \cellcolor{tabsecond}0.065  \\
3D Gaussian + ActiveNeRF & \cellcolor{tabsecond}27.326 &  \cellcolor{tabthird}0.912 &  \cellcolor{tabthird}0.076  \\
\hline
Plenoxels + Ours          &                      24.212 &                      0.878 &                      0.139  \\
3D Gaussian + Ours       &  \cellcolor{tabfirst}29.094 &  \cellcolor{tabfirst}0.938 &  \cellcolor{tabfirst}0.053 

    \end{tabular}
    }
\end{minipage}
\end{table}
\begin{table}
    \centering
    \caption{{\bf Active View Selections on Blender Dataset with only 10 views} Our view selection algorithm could select necessary views when the number of training views is extremely limited. }\label{tab:blender10}
    \resizebox{0.48\textwidth}{!}{
    \begin{tabular}{l|ccc}
    Method & PSNR	 $\uparrow$ &	SSIM  $\uparrow$	&	LPIPS $\downarrow$\\
    \hline
    
Nerfacto + Random        &                      16.430 &                      0.809 &                      0.254  \\
Nerfacto + BayesRays     &                      14.950 &                      0.798 &                      0.282  \\
Plenoxels + Random        &                      19.950 &                      0.812 &                      0.233  \\
Plenoxels + ActiveNeRF    &                      19.770 &                      0.804 &                      0.210  \\
3D Gaussian + Random     &  \cellcolor{tabthird}22.493 &  \cellcolor{tabthird}0.873 &  \cellcolor{tabthird}0.112  \\
3D Gaussian + ActiveNeRF & \cellcolor{tabsecond}22.979 & \cellcolor{tabsecond}0.876 & \cellcolor{tabsecond}0.111  \\
\hline
Plenoxels + Ours          &                      20.670 &                      0.824 &                      0.205  \\
3D Gaussian + Ours       &  \cellcolor{tabfirst}23.681 &  \cellcolor{tabfirst}0.883 &  \cellcolor{tabfirst}0.102 

    \end{tabular}
    }
\end{table}

\begin{table}
\begin{minipage}{0.47\textwidth}
    \centering
    \caption{{\bf Active View Selections on Mip-NeRF360 Dataset} We compare our method on real-world dataset Mip360. $\dag$: batch active view selection setting. Our method outperformed previous state-of-the-art with a clear margin.} \label{tab:mip360}
    \resizebox{\textwidth}{!}{
    \begin{tabular}{l|ccc}
    Method & PSNR	 $\uparrow$ &	SSIM  $\uparrow$	&	LPIPS $\downarrow$\\
    \hline
    
3D Gaussian + Random         &                      17.914 &                      0.564 &                      0.430  \\
3D Gaussian + Random\dag     &  \cellcolor{tabthird}19.542 &  \cellcolor{tabthird}0.568 &  \cellcolor{tabthird}0.376  \\
3D Gaussian + ActiveNeRF     &                      17.889 &                      0.533 &                      0.414  \\
3D Gaussian + ActiveNeRF\dag &                      18.303 &                      0.539 &                      0.406  \\
\hline
3D Gaussian + Ours           & \cellcolor{tabsecond}20.351 & \cellcolor{tabsecond}0.601 &  \cellcolor{tabfirst}0.361  \\
3D Gaussian + Ours\dag       &  \cellcolor{tabfirst}20.568 &  \cellcolor{tabfirst}0.608 & \cellcolor{tabsecond}0.365

    \end{tabular}
    }
\end{minipage}
    ~
\begin{minipage}{0.47\textwidth}
    \centering
    \caption{{\bf Scene Coverage on Gibson and MP3D dataset}
    Our method outperformed previous methods by a large margin when evaluating the coverage metric. 
    Numbers for other methods are taken from Active Nerual Mapping~\cite{yan2023active-neural-mapping}.
    }\label{tab:mapping}
    \resizebox{\textwidth}{!}{
    \begin{tabular}{lcccc}
    \toprule
\multirow{2}{*}{Method} & \multicolumn{2}{c}{Gibson} & \multicolumn{2}{c}{MP3D} \\ 
            \cmidrule(lr){2-3} \cmidrule(lr){4-5}
            & Comp. (\%) $\uparrow$  & Comp. (cm) $\downarrow$ & Comp. (\%)$\uparrow$   & Comp. (cm)  $\downarrow$\\
            \toprule
            Random &   45.80     &   34.48    &    45.67   &   26.53   \\
            FBE    & \cellcolor{tabthird}  68.91     & \cellcolor{tabthird}  14.42     &   71.18     &  9.78    \\
            UPEN   &   63.30     &   21.09     &   69.06     &   10.60     \\
            OccAnt &  61.88      &  23.25     &   \cellcolor{tabthird} 71.72    & \cellcolor{tabthird}   9.40   \\
            Active Neural Mapping & \cellcolor{tabsecond} 80.45 & \cellcolor{tabsecond} 7.44 & \cellcolor{tabsecond} 73.15 & \cellcolor{tabsecond} 9.11 \\
            \midrule
            Ours  &   \cellcolor{tabfirst} 92.89  & \cellcolor{tabfirst} 5.64 & \cellcolor{tabfirst} 89.41 & \cellcolor{tabfirst} 2.91    \\
            \bottomrule
\end{tabular}
    }
\end{minipage}
\end{table}

\noindent
The quantitative results of active view selections can be found in Table.~\ref{tab:blender20}, Table.~\ref{tab:blender-batch} and Table.~\ref{tab:mip360}. As can be seen, our method achieved better results across different metrics and datasets. We also compare our method qualitatively on the Blender and Mip-NeRF360 datasets in Fig.~\ref{fig:mip} and Fig.~\ref{fig:qual-blender}.
Our method demonstrates superior results in both single-view selection and batch-view selection.
Moreover, our method achieved better performance improvements compared to the random backbone compared to BayesRays. This further renders the importance of using EIG rather than uncertainty as the metric for view selection.
Furthermore, we experimented with our model with the challenging ten-view selection task on the Blender Dataset. Each model is initialized with the same random seed and two uniform initial views. Each new view is added after every 100 epochs till the model has ten training views. The quantitative and qualitative results are in Table.~\ref{tab:blender10} and Fig.~\ref{fig:qual-blender}. Again, our method selects necessary views given the extremely limited observations and preserves fine details of reconstructed objects. 
\subsection{Active Mapping}\label{sec:mapping-exp}
\label{sec:active-map}
We experimented with our active mapping system on room-scale scene dataset Matterport3D~\cite{Matterport3D} and Gibson~\cite{xia2018gibson} through the Habitat Simulator~\cite{habitat19iccv, ramakrishnan2021hm3d}.
Our active mapping system takes posed RGB-D images as input and uses the same configurations for the simulator as previous approaches~\cite{yan2023active-neural-mapping, upen}.
We turn our 3D Gaussian representation into a point cloud using only the mean $\mathbf{\mu} \in R^3$ of each 3D Gaussian as the previous evaluation benchmark is designed for point cloud or 3D mesh. 
We use the following evaluation metric to evaluate the quality of our active mapping system.

\noindent\textbf{Comp.}~The completeness metric quantifies the completeness of the active exploration algorithm in 3D space by computing the per-point nearest distance between points sampled from the ground truth mesh and the final reconstruction of the scene. We calculate both the percentage of points within a 5cm threshold (Comp. (\%)) and the average nearest distance (Comp. ($cm$)).

\noindent\textbf{PSNR \& Depth MAE}~We uniformly sample 1000 navigable poses in the scene and render color and depth images at the sampled poses and compute average PSNR and Mean Absolute Error (MAE) for rendered color and depth images, respectively.
\begin{figure*}[!t]
\begin{tabular}{ccc}
Ground Truth & Active-INR & Ours   \\
\includegraphics[width=0.32\linewidth]{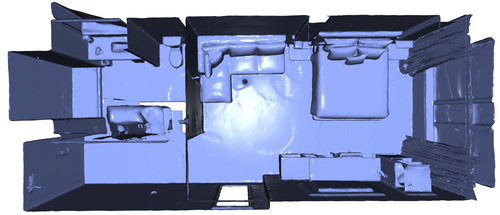} & \includegraphics[width=0.32\linewidth]{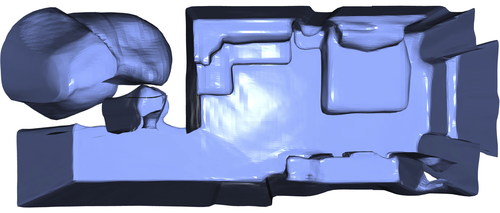} & \includegraphics[width=0.32\linewidth]{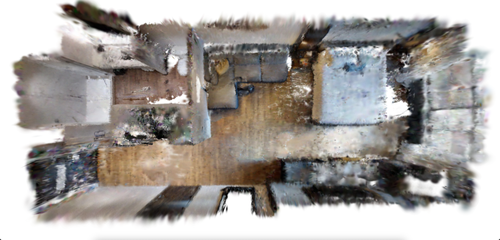} \\
\includegraphics[width=0.32\linewidth]{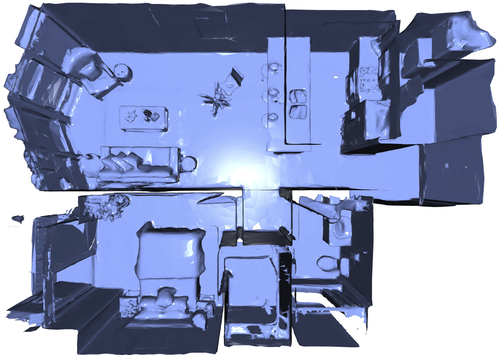} & \includegraphics[width=0.32\linewidth]{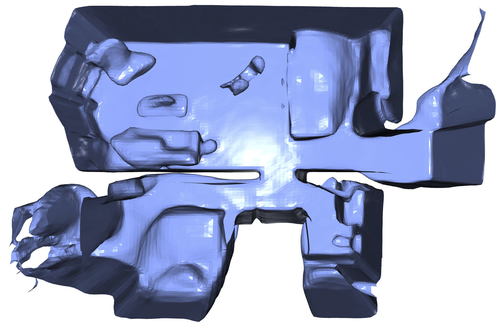} & \includegraphics[width=0.32\linewidth]{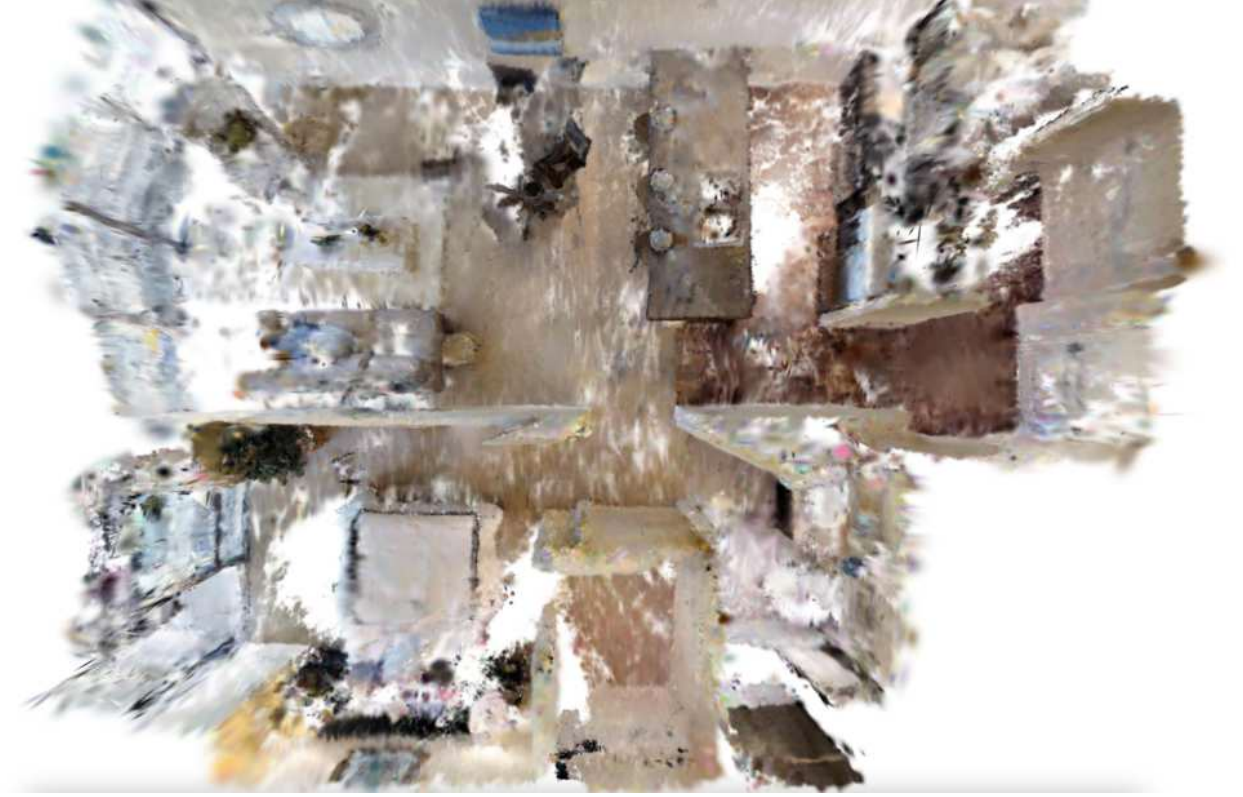} \\
\end{tabular}
	\caption{{\bf Qualitative Comparisons on Active Mapping} We compare our method against the ground truth mesh and Active Neural Mapping. Our method exhibited better details and coverage.
Our method can successfully produce detailed and high-fidelity reconstruction for the rooms compared to the previous method. 
}
\label{fig:mp3d}
\vspace{-3mm}
\end{figure*}

\begin{table}[t]
\begin{minipage}{0.47\textwidth}
    \centering
    \caption{{\bf Evaluation for Render Quality on Gibson and MP3D} We also compare our method with previous method
    on the rendering quality for both color and depth reconstruction.}\label{tab:map-render}
    \label{table:render}
    \resizebox{\textwidth}{!}{
    \begin{tabular}{lcccc}
    \toprule
\multirow{2}{*}{Method} & \multicolumn{2}{c}{Gibson} & \multicolumn{2}{c}{MP3D} \\ 
            \cmidrule(lr){2-3} \cmidrule(lr){4-5}
            & PSNR $\uparrow$  & Dpeth MAE (m) $\downarrow$ & PSNR $\uparrow$  & Depth MAE (m) $\downarrow$ \\
            \toprule
            FBE &   19.35    &  \cellcolor{tabsecond} 0.1751   & 16.68  & \cellcolor{tabthird} 0.3627   \\
            UPEN    &  \cellcolor{tabsecond} 21.13 & \cellcolor{tabthird} 0.1893 & \cellcolor{tabsecond}  18.40    &  \cellcolor{tabsecond} 0.2690    \\
            Active Neural Mapping   & \cellcolor{tabthird} 19.64 &  0.2943   & \cellcolor{tabthird}  17.16  & 0.4621     \\
            \midrule
            Ours  &  \cellcolor{tabfirst} 22.58  & \cellcolor{tabfirst} 0.0924 & \cellcolor{tabfirst} 19.96 & \cellcolor{tabfirst} 0.1667 \\
            \bottomrule
\end{tabular}
    }
\end{minipage}
    ~
\begin{minipage}{0.47 \textwidth}
    \centering
    \caption{{\bf Uncertainty Estimation on LF Dataset} Numbers are AUSE, the lower the better. The best results are highlighted in red, and the second-best results are in orange color. }\label{tab:ucnern-gaussian}
    \resizebox{\textwidth}{!}{
    \begin{tabular}{l|ccccc}
   Method & Statue $\downarrow$	&	Africa	 $\downarrow$&	Torch $\downarrow$	&	Basket $\downarrow$ & Average  $\downarrow$\\
    \hline
CF-NeRF & 0.54 & \cellcolor{tabthird}0.34 & 0.50 &  \cellcolor{tabfirst}0.14 & 0.38 \\
\hline
 ActiveNeRF    &  \cellcolor{tabthird} 0.40 &  0.39 &  \cellcolor{tabfirst} 0.21 &  0.34 &  \cellcolor{tabthird}0.33 \\
 \hline
 BayesRays    &  \cellcolor{tabfirst} 0.17 &  \cellcolor{tabsecond} 0.27 &  \cellcolor{tabsecond} 0.22 & \cellcolor{tabthird} 0.28 &  \cellcolor{tabsecond}0.23 \\
\hline
 Ours    &  \cellcolor{tabsecond}0.21 &  \cellcolor{tabfirst}0.26 &  \cellcolor{tabthird}0.24 & \cellcolor{tabsecond}0.18 &  \cellcolor{tabfirst}0.22
    \end{tabular}
    }
\end{minipage}
\end{table}
\vspace{-5mm}

We compare with previous state-of-the-art in Table.~\ref{tab:mapping} and Figure.~\ref{fig:mp3d}. 
Our active mapping system outperformed previous methods by a large margin on both geometry and rendering quality.
Notably, we are the first active mapping system to capture fine-grained details and high-quality textures, thanks to our view selection objectives, which aim to improve rendering quality when measuring the EIG. 
In Figure \ref{fig:mp3d}, we demonstrate the qualitative comparison between our method and Active-INR \cite{yan2023active-neural-mapping}. 
In Table \ref{tab:mapping}, our method outperforms Active-INR by a large margin in Comp. metric, showing the superiority of the active mapping algorithm. To compare the rendering quality with previous methods, we train a 3D Gaussian Splatting model using the trajectory from UPEN~\cite{upen} and Active-INR~\cite{yan2023active-neural-mapping} and evaluate the render quality at the same holdout set. The result is summarized in Table \ref{table:render}. Our method efficiently selects the most informative path and outperforms other methods when using the same backbone model. We also notice that UPEN is worse than Active Neural Mapping in Table \ref{tab:mapping} but better than the other in \ref{table:render} when evaluated using the 3D Gaussian Splatting model. We ascribe this gap in UPEN to the difference in the scene representations and the evaluation protocol.
 
\subsection{Uncertainty Quantification}
\label{sec:exp_uncern}
\begin{figure}[t]
\vspace{-3mm}
 \resizebox{\textwidth}{!}{
\vspace{-7mm}
\begin{tabular}{cc|cc|cc}
\hline
\hline
\multicolumn{2}{c|}{CF-NeRF} & \multicolumn{2}{c|}{BayesRays} & \multicolumn{2}{c}{Ours}  \\
\hline
Uncertainty & Depth Err. &  Uncertainty & Depth Err. &  Uncertainty & Depth Err. \\
\hline
\hline
\includegraphics[width=.3\linewidth]{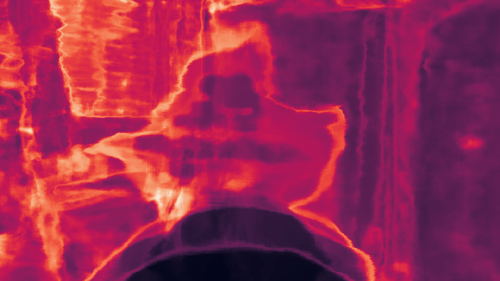} & 
\includegraphics[width=.3\linewidth]{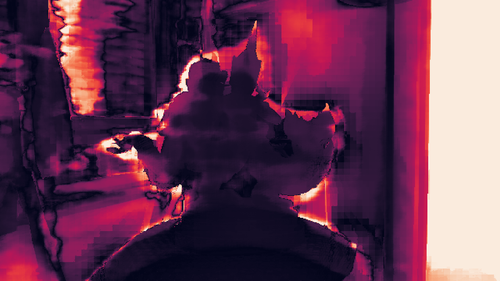} & 
\includegraphics[width=.3\linewidth]{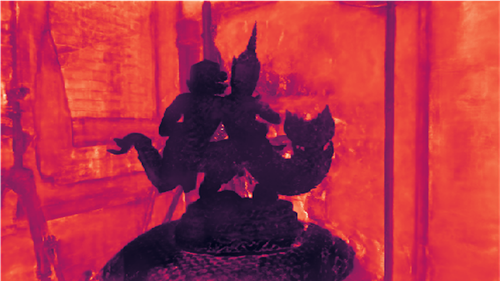} & 
\includegraphics[width=.3\linewidth]{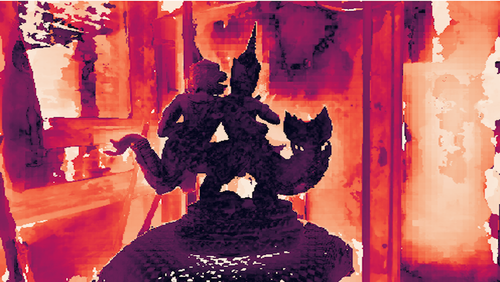} & 
\includegraphics[width=.3\linewidth]{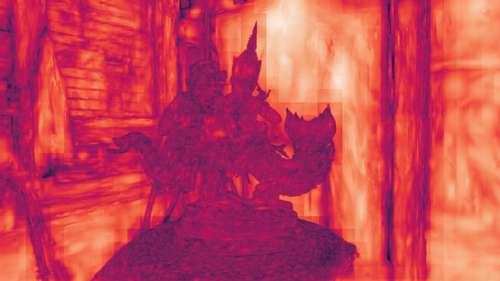} & 
\includegraphics[width=.3\linewidth]{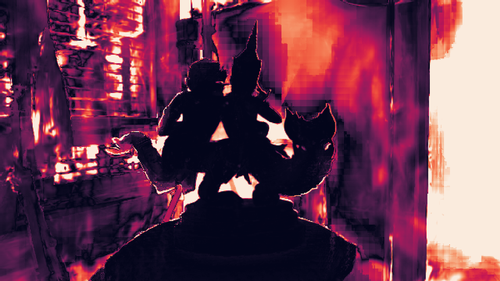} \\
\includegraphics[width=.3\linewidth]{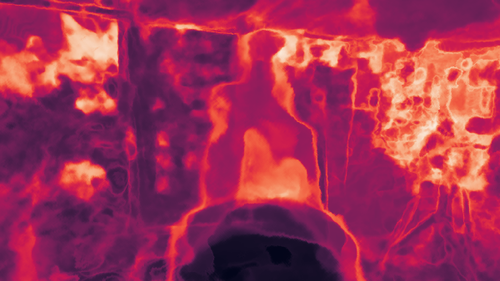} & 
\includegraphics[width=.3\linewidth]{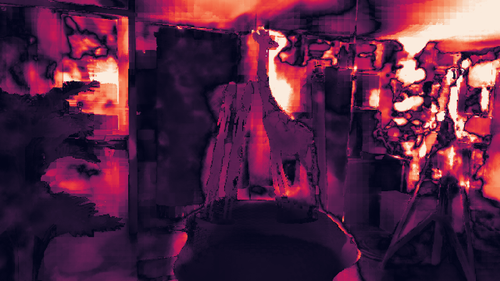} & 
\includegraphics[width=.3\linewidth]{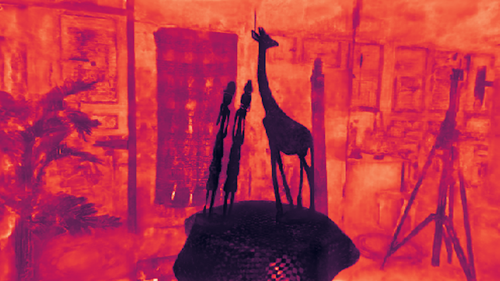} & 
\includegraphics[width=.3\linewidth]{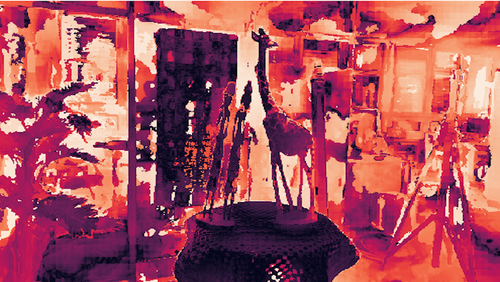} &  \includegraphics[width=.3\linewidth]{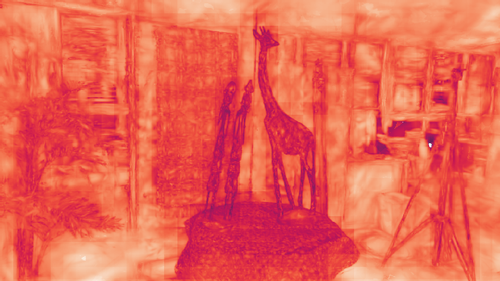} & 
\includegraphics[width=.3\linewidth]{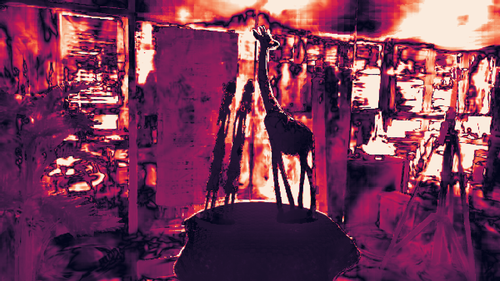} \\
\includegraphics[width=.3\linewidth]{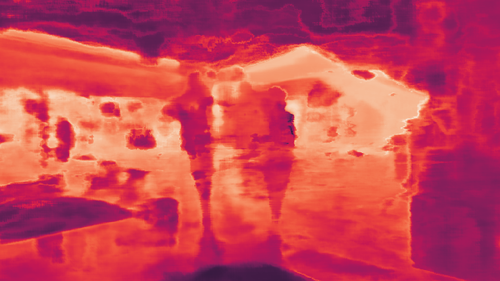} & 
\includegraphics[width=.3\linewidth]{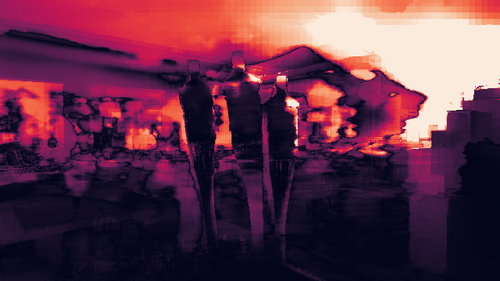} & 
\includegraphics[width=.3\linewidth]{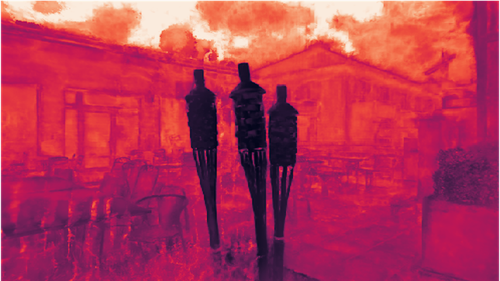} & 
\includegraphics[width=.3\linewidth]{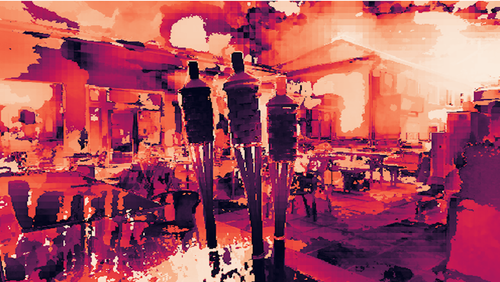} &  
\includegraphics[width=.3\linewidth]{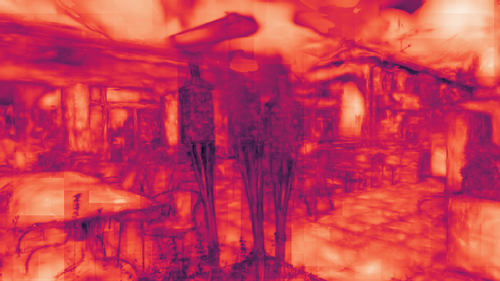} & 
\includegraphics[width=.3\linewidth]{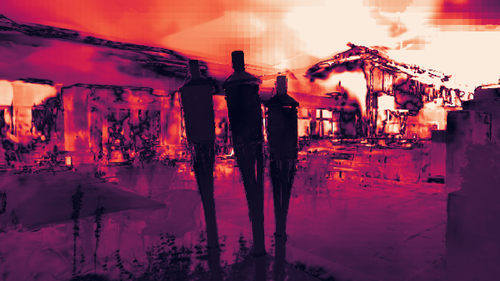}\\
\hline
\hline
\end{tabular}
}
\vspace{-2mm}
\caption{{\bf Visualizations for Uncertainty on Depth Prediction the LF Dataset}
We compare the per-pixel uncertainty maps from CF-NeRF~\cite{CF-NeRF}, BayesRays~\cite{goli2023}, and ours. Our per-pixel uncertainty maps demonstrate a strong correlation with the actual depth error.
\vspace{-6mm}
}
 \label{fig:uncern}
\end{figure}

As discussed in Sec.~\ref{sec:pixel-uncern}, our model can be extended to compute pixel-wise uncertainties on training views. 
Following previous methods on uncertainty estimation~\cite{CF-NeRF, shen2021snerf}, we evaluate our method on the Light Field~(LF) Dataset~\cite{lf-dataset} using the Area Under Sparsification Error (AUSE) metric. The pixels are filtered twice, once using the absolute error with ground truth depth and once using the uncertainty. The difference in the mean absolute error on the remaining pixels between the two sparsification processes produces two different error curvatures, where the area between those two curvatures is the AUSE, which evaluates the correlation between uncertainties and the predicted error. 
A low AUSE indicates our model is confident in the correctly estimated depths and could predict a high uncertainty in the regions where we are likely to have larger errors.
As seen in Table~\ref{tab:ucnern-gaussian} and Fig.~\ref{fig:uncern}, our model exhibited a better correlation between our uncertainty estimation and depth error than the previous state-of-the-art CF-NeRF~\cite{CF-NeRF}. More quantitative results and visualizations on uncertainty estimation can be found in our supplementary materials. 

\vspace{-3mm}
\section{Conclusion and Limitations} \label{sec:conclusion}
\vspace{-2mm}
We presented FisherRF, a novel method for active view selection, active mapping, and uncertainty quantification in Radiance Fields. 
Leveraging Fisher Information, our method provides an efficient and effective means to quantify the observed information of Radiance Field models. The flexibility of our approach allows it to be applied to various model parametrizations, including 3D Gaussian Splatting and Plenoxels.
Our extensive evaluation of active view selection, active mapping, and uncertainty quantification has consistently shown superior performance compared to existing methods and heuristic baselines. These results highlight the potential of our approach to significantly enhance the quality and efficiency of image rendering and reconstruction tasks with limited viewpoints.
However, our method is limited to static scenes in a confined scenario. Reconstructing dynamically changing the Radiance Field and quantifying its Fisher Information is still an open problem.
More work could be done to overcome the limitations and extend the proposed method to more challenging settings. 

{\small
\paragraph{Acknowledgements}
The authors gratefully appreciate support through the following grants: NSF FRR 2220868, NSF IIS-RI 2212433, NSF TRIPODS 1934960, NSF CPS 2038873.
The authors thank Pratik Chaudhari for the insightful discussion and Yinshuang Xu for proofreading the drafts.
}

\bibliographystyle{splncs04}
\bibliography{main}
\end{document}